%% file: ms.tex
\documentclass[10pt,twocolumn,letterpaper]{article}
\usepackage{cvpr}
\usepackage{times}
\usepackage[noend]{algpseudocode}
\usepackage{algorithm}
\usepackage{algorithmicx}
\usepackage{amsmath}
\usepackage{amssymb}
\usepackage{bm}
\usepackage{booktabs}
\usepackage{booktabs}
\usepackage{comment}
\usepackage{enumitem}
\usepackage{graphicx}
\usepackage{listings}
\usepackage{placeins}
\usepackage[backgroundcolor=white,linecolor=red,bordercolor=none,textwidth=3em,textsize=tiny]{todonotes}
\usepackage[pagebackref=true,breaklinks=true,letterpaper=true,colorlinks,bookmarks=false]{hyperref}
\usepackage{cleveref}

\newcommand{\bbM}{\bm{M}}
\newcommand{\bbm}{\bm{m}}

\newcommand{\bw}{\bm{w}}
\newcommand{\bx}{\bm{x}}
\newcommand{\by}{\bm{y}}

\newcommand{\bmx}{\mathbf{X}}
\newcommand{\bmw}{\mathbf{W}}

\cvprfinalcopy

\ifcvprfinal\fi

\title{There and Back Again: Revisiting Backpropagation Saliency Methods}
\author{Sylvestre-Alvise Rebuffi \thanks{indicates equal contribution} \hspace{2em} Ruth Fong\footnotemark[1] \hspace{2em} Xu Ji\footnotemark[1] \hspace{2em} Andrea Vedaldi\\
Visual Geometry Group, University of Oxford\\
{\tt\small \{srebuffi, ruthfong, xuji, vedaldi\}@robots.ox.ac.uk}
}

\begin{document}
\maketitle
\thispagestyle{empty}
\begin{abstract}
Saliency methods seek to explain the predictions of a model by producing an importance map across each input sample. A popular class of such methods is based on backpropagating a signal and analyzing the resulting gradient. Despite much research on such methods, relatively little work has been done to clarify the differences between such methods as well as the desiderata of these techniques. Thus, there is a need for rigorously understanding the relationships between different methods as well as their failure modes. In this work, we conduct a thorough analysis of backpropagation-based saliency methods and propose a single framework under which several such methods can be unified. As a result of our study, we make three additional contributions. First, we use our framework to propose NormGrad, a novel saliency method based on the spatial contribution of gradients of convolutional weights. Second, we combine saliency maps at different layers to test the ability of saliency methods to extract complementary information at different network levels (e.g.~trading off spatial resolution and distinctiveness) and we explain why some methods fail at specific layers (e.g., Grad-CAM anywhere besides the last convolutional layer). Third, we introduce a class-sensitivity metric and a meta-learning inspired paradigm applicable to any saliency method for improving sensitivity to the output class being explained.
\end{abstract}

\input{intro}

\input{related}

\input{method}

\input{experiments}

\input{conclusions}

{\small\bibliographystyle{ieee_fullname}\bibliography{ms}}
\appendix\input{supps}
\end{document}

%% file: intro.tex
\section{Introduction}\label{s:introduction}
The adoption of deep learning methods by high-risk applications, such as healthcare and automated driving, gives rise to a need for tools that help machine learning practitioners understand model behavior.
Given the highly-parameterized, opaque nature of deep neural networks, developing such methods is non-trivial, and there are many possible approaches. In the basic case, even the predictions of the model itself, either unaltered or after being distilled into a simpler function~\cite{frosst2017distilling, ba2014deep}, can be used to shed light on its behaviour.

Saliency is the specific branch of interpretability concerned with determining not what the behaviour of a model is for whole input samples, but which parts of samples contribute the most to that behaviour. Thus by definition, determining saliency - or attribution - necessarily involves reversing the model's inference process in some manner~\cite{mahendran16salient}. 
Propagating a signal from the output layer of a neural network model back to the input layer is one way of explicitly achieving this.

The number of diverse works based on using signal backpropagation for interpretability in computer vision~\cite{zeiler2014visualizing, selvaraju17gradcam, bach2015pixel, zhang2016excitation} is testimony to the power of this simple principle. Typically, these techniques produce a heatmap for any given input image that ranks its pixels according to some metric of importance for the model's decision. Inspired by the work of~\cite{Adebayo2018Sanity}, we propose to delve deeper into such methods by discussing some of the similarities, differences and potential improvements that can be illustrated. 




We begin with presenting a framework that unifies several backpropagation-based saliency methods by dividing the process of generating a saliency map into two phases: \emph{extraction} of the contribution of the gradient of network parameters at each spatial location in a particular network layer, and \emph{aggregation} of such spatial information into a 2D heatmap. GradCAM~\cite{selvaraju17gradcam}, linear approximation~\cite{kindermans2016investigating} and gradient~\cite{simonyan14deep} can all be cast as such processes.
Noting that no appropriate technique has yet been proposed for properly aggregating contributions from convolutional layers, we introduce NormGrad, which uses the Frobenius norm for aggregation. We introduce identity layers to allow for the generation of saliency heatmaps at all spatially-grounded layers in the network (i.e. even after ReLU), since NormGrad computes saliency given a parameterised network layer. 



We conduct a thorough analysis of backpropagation-based saliency methods in general, with evaluation based on utilising saliency heatmaps for weak localisation. Notably, we conduct an investigation into simple techniques for combining saliency maps taken from different network layers - in contrast to the popular practice of computing maps at the input layer~\cite{simonyan14deep} - and find that using a weighted average of maps from all layers consistently improves performance for several saliency methods, compared to taking the single best layer. However, not all layers are equally important, as we discover that models optimised on datasets such as ImageNet~\cite{russakovsky2015imagenet} and PASCAL VOC~\cite{everingham2015pascal} learn features that become increasingly \emph{class insensitive} closer towards the input layer. This provides an explanation for why Grad-CAM~\cite{selvaraju17gradcam} produces unmeaningful saliency heatmaps at certain layers earlier than the last convolutional layer, as the sensitivity of the gradient to class across spatial locations is eliminated by Grad-CAM's spatial gradient averaging, meaning such layers are devoid of class sensitive signals from which to form saliency heatmaps.


Finally, building off~\cite{mahendran16salient,Adebayo2018Sanity}, we introduce a novel metric for quantifying the class sensitivity of a saliency method, and present a meta-learning inspired paradigm that increases the class sensitivity of any method by adding an inner SGD step into the computation of the saliency heatmap.

%% file: related.tex
\section{Related work}\label{s:related}

\paragraph{Saliency methods.}
Our work focuses on backpropagation-based saliency methods; these techniques are computationally efficient as they only require one forward and backward pass through a network.
One of the earliest methods was~\cite{simonyan14deep} which visualised the gradient at the input with respect to an output class being explained.
Several authors have since proposed adaptations in order to improve the heatmap's visual quality.
These include modifying the ReLU derivatives~(Deconvnet~\cite{zeiler2014visualizing}, guided backprop~\cite{springenberg2014striving}) and averaging over randomly perturbed inputs~(SmoothGrad~\cite{smilkov2017smoothgrad}) to produce masks with reduced noise.
Several works have explored visualizing saliency at intermediate layers by combining information from activations and gradients, notably CAM~\cite{zhou2016learning}, Grad-CAM~\cite{selvaraju17gradcam}, and linear approximation (a.k.a. gradient $\times$ input)~\cite{kindermans2016investigating}.
Conservation-preserving methods (Excitation Backprop~\cite{zhang2016excitation}, LRP~\cite{bach2015pixel}, and DeepLIFT~\cite{shrikumar2017learning}) modify the backward functions of network layers in order to ``preserve'' an attribution signal such that it sums to one at any point in the network.
Reference-based methods average over attributions from multiple interpolations~\cite{sundararajan2017axiomatic} between the input and a non-informative (e.g. black) reference input or use a single reference input with which to compare a backpropagated attribution signal~\cite{shrikumar2017learning}.

Although we focus on backpropagation-based methods, another class of methods studies the effects that perturbations on the input induce on the output.
\cite{zeiler2014visualizing} and~\cite{Petsiuk2018rise} generate saliency maps by weighting input occlusion patterns by the induced changes in model output.
\cite{fong17interpretable, fong19,kapishnikov2019xrai} learn for saliency maps that maximally impact the model, and ~\cite{dabkowski2017real} trains a model to predict effective maps.
LIME~\cite{ribeiro2016should} learns linear weights that correspond to the effect of including or excluding (via perturbation) different image patches in an image.
Perturbation-based approaches have also been used to perform object localisation~\cite{singh2017hide, wei2017object, wang2017fast}.

\paragraph{Assessing and unifying saliency methods.}
A few works have studied if saliency methods have certain desired sensitivities (e.g., to specific model weights~\cite{Adebayo2018Sanity} or the output class being explained~\cite{mahendran16salient}) and if they are invariant to unmeaningful factors (e.g., constant shift in input intensity~\cite{kindermans2019reliability}).
\cite{mahendran16salient} showed that gradient~\cite{simonyan14deep}, deconvnet~\cite{zeiler2014visualizing}, and guided backprop~\cite{springenberg2014striving} tend to produce class insensitive heatmaps.
\cite{Adebayo2018Sanity} introduced sanity check metrics that measure how sensitive a saliency method is to model weights by reporting the correlation between a saliency map on a trained model vs. a partially randomized model.

Other works quantify the utility of saliency maps for weak localization~\cite{zhang2016excitation,fong17interpretable} and for impacting model predictions.
\cite{zhang2016excitation} introduced Pointing Game, which measures the correlation between the maximal point extracted from a saliency map with pixel-level semantic labels.
\cite{fong17interpretable} extracts bounding boxes from saliency maps and measures their agreement with ground truth bounding boxes. \cite{yang2019bench} evaluates attribution methods using relative feature importance. \cite{oramas2019visual} proposes a dataset designed for measuring visual explanation accuracy.
\cite{bach2015pixel,Petsiuk2018rise,kapishnikov2019xrai} present variants of a perturbation-based evaluation metric that measures the impact of perturbing (or unperturbing for~\cite{kapishnikov2019xrai}) image patches in order of importance as given by a saliency map. 
However, these perturbed images are outside the training domain;~\cite{hooker2018evaluating} mitigates this by measuring the performance of classifiers re-trained on perturbed images (i.e., with 20\% of pixels perturbed). 

To our knowledge, the only work that has been done to unify saliency methods focuses primarily on ``invasive'' techniques that change backpropagation rules.
The $\alpha$-LRP variant~\cite{bach2015pixel} and Excitation Backprop~\cite{zhang2016excitation} share the backpropagation rule, and DeepLIFT~\cite{shrikumar2017learning} is equivalent to LRP when $0$ is used as the reference activation throughout a network.
\cite{lundberg2017unified} unifies a few methods (e.g., LIME~\cite{ribeiro2016should}, LRP~\cite{bach2015pixel}, DeepLIFT~\cite{shrikumar2017learning}) under the framework of additive feature attribution.

%% file: method.tex
\section{Method}\label{s:method}

\paragraph{Preliminaries.} Consider a training set $\mathcal{D}$ of pairs $(\bx,\by)$ where $\bx \in \mathbb{R}^{3\times H\times W}$ are (color) images and $\by \in \{1,\dots,C\}$ their labels.
Furthermore, let $\by = \Phi_\theta(\bx)$ be the output of a neural network model whose parameters $\theta$ are optimized using the cross-entropy loss $\ell$ to predict labels from images.

\subsection{Extract \& Aggregate framework}

In most methods, the saliency map is obtained as a function of the network activations, computed in a forward pass, and information propagated from the output of the network back to its input using the backpropagation algorithm.
While some methods modify backpropagation in some way, here we are interested in those, such as gradient, linear approximation, and all variants of our proposed NormGrad saliency method, that do not.

In order to explain these ``non-invasive'' methods, we suggest that their saliency maps can be interpreted as a measure of how much the corresponding  pixels contribute to changing the model parameters during a training step.
We then propose a principled two-phase framework capturing this idea.
In the extraction phase, a method isolates the contribution to the gradient from each spatial location.
We use the fact that the gradient of spatially shared weights can be written as the sum over spatial locations of a function of the activation gradients and input features.
In the aggregation phase, each spatial summand is converted into a scalar using an aggregation function, thus resulting in a saliency map.





\input{figures/procedure.tex}

\subsubsection{Phase 1: Extract}\label{s:extract}
We first choose a target layer in the network at which we plan to compute a saliency map.
Assuming that the network is a simple chain\footnote{Other topologies are treated in the same manner, but the notation is more complex.}, we can write $L = \ell \circ \Phi = h \circ k_{\bw} \circ q$, where $k_{\bw}$ is the target layer parameterised by $\bw$, $h$ is the composition of all layers that follow it (including loss $\ell$), and $q$ is the composition of layers that precede it.
Then, $\bx^{in} = q(\bx) \in \mathbb{R}^{K\times H\times W}$ denotes the input to the target layer, and its output is given by $\bx^{out} = k_{\bw}(\bx^{in}) \in \mathbb{R}^{K'\times H'\times W'}$.

\paragraph{Convolutional layers with general filter shapes.}
For convolutions with an $N \times N$ kernel size, we can re-write the convolution using the matrix form:
\begin{equation}
 \bmx^{out} = \bmw \bmx^{in}_{N \times N}
\end{equation}
where $\bmx^{out} \in \mathbb{R}^{K' \times HW}$ and $\bmw \in \mathbb{R}^{K'\times N^2K}$ are the output and filter tensors reshaped as matrices and $\bmx^{in}_{N \times N} \in \mathbb{R}^{N^2K \times HW}$ is a matrix whose column $\bx^{in}_{u, N \times N} \in \mathbb{R}^{N^2K}$ contains the unfolded patches at location $u$ of the input tensor to which filters are applied.\footnote{This operation is called \texttt{im2row} in MATLAB or \texttt{unfold} in PyTorch.}
Then, the gradient w.r.t.~the filter weights $W$ is given by
\begin{equation}
\frac{dL}{d\bmw}
=
\sum_{u\in\Omega}
\frac{d}{d\bmw}
\left\langle
\bm{g}^{out}_{u},
\bmw \bx^{in}_{u, N \times N}
\right\rangle
=
\sum_{u\in\Omega}
\bm{g}^{out}_{u} {\bx^{in}_{u, N \times N}}^\top,
\end{equation}
where $\bm{g}^{out}_{u} = dh/d\bx^{out}_u$ is the gradient of the ``head'' of the network.
Thuis, for the convolutional layer case, the spatial summand is an outer product of two vectors; thus, the spatial contribution at each location to the gradient of the weights is a matrix of size $K' \times N^2K$.

\paragraph{Other layer types.}
Besides convolutional layers, bias and scaling layers also share their parameters spatially.
In modern architectures, these are typically used in batch normalization layers~\cite{ioffe2015batch}. 
We denote $\bm{b}, \bm{\alpha} \in \mathbb{R}^K$ as the parameters for the bias and scaling layers respectively.
They are defined respectively as follows:
$$
x^{out}_{ku}
=
x^{in}_{ku} + b_k,
\quad\quad\quad
x^{out}_{ku}
=
\alpha_k x^{in}_{ku}.
$$
Then, the gradients w.r.t. parameters are given by
\begin{equation}
\frac{dL}{d\bm{b}}
=
\sum_{u\in\Omega}
\bm{g}^{out}_{u},
\quad\quad\quad
\frac{dL}{d\bm{\alpha}}
=
\sum_{u\in\Omega}
\bm{g}^{out}_{u} \odot {\bx^{in}_{u}}.
\end{equation}
where $\odot$ is the elementwise product. For these two types of layer, the spatial summand is a vector of size $K$, the number of channels. \Cref{t:layers} summarizes the spatial contributions to the gradients for the different layers.

\begin{table}
\setlength\tabcolsep{3pt}
\begin{center}
\begin{tabular}{|r|c|l|}
\hline
Layer & Spatial contribution & Size at each location \\
\hline
Bias & $\bm{g}^{out}_{u}$ & vector: $K'$ \\
Scaling & $  \bm{g}^{out}_{u} \odot \bx^{in}_{u}$ & vector: $K'$ \\
Conv $N \times N$ & ${\bm{g}^{out}_u}{\bx^{in}_{u,N \times N}}^\top$ & matrix: $K' \times N^2K$ \\
\hline
\end{tabular}
\end{center}
\caption{Formulae and sizes of the spatial contributions to the gradient of the weights for layers with spatially shared parameters. $\odot$ denotes the elementwise product and $\bx^{in}_{u,N \times N}$ is the $N^2K$ vector obtained by unfolding the $N \times N$ patch around the target location.}
\label{t:layers}
\end{table}

\subsubsection{Virtual identity layer}~\label{s:virt_ident}
So far, we have only extracted spatial gradient contributions for layers that share parameters spatially.
We will now extend our summand extraction technique to any location within a CNN by allowing the insertion of a virtual identity layer at the target location.
This layer is a conceptual construction that we introduce to derive in a rigorous and unified manner the equations employed by various methods to compute saliency.

We are motivated by the following question:
if we were to add an identity operator at a target location, how should this operator's parameters be changed?
A virtual identity layer is a layer which shares its parameters spatially and is set to the identity.
Hence, it could be any of the layer from~\cref{t:layers}, like a bias or a scaling layer; then, $b_k = 0 $ or $c_k = 1$ respectively for $k \in \{1,\dots,K\}$.
It could also be a $N \times N$ convolutional layer with filter bank $\bmw \in \mathbb{R}^{K \times N^2K}$ such that $w_{kk'ij} = \delta_{k,k'}\delta_{i,0}\delta_{j,0}$, where $\delta_{a,b}$ is the Kronecker delta function\footnote{Kronecker delta function: $\delta_{a,b} = 1 \text{ if } a = b$; otherwise, $\delta_{a,b} = 0$.}, for $(i,j) \in \{-\frac{N-1}{2},\dots,\frac{N-1}{2}\}$.

\begin{figure}[!htb]
    \centering
    \includegraphics[width=0.9\linewidth]{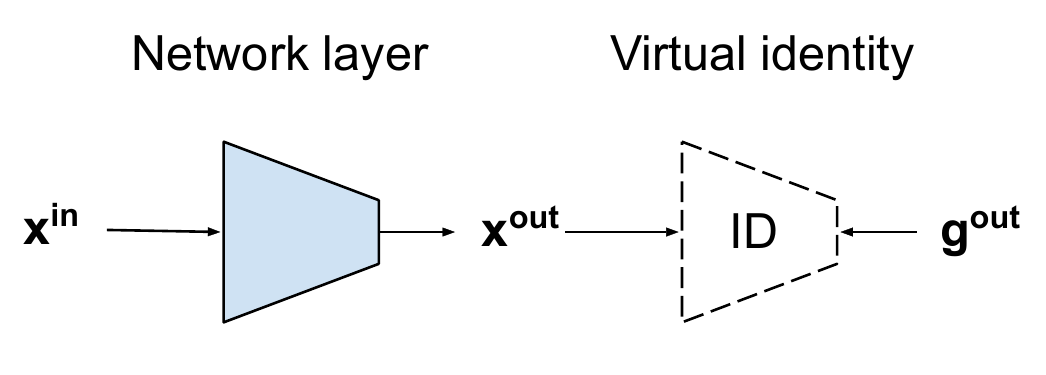}
    \caption{\textbf{Virtual identity.} Here, we visualize inserting an identity after a specific layer in the network for saliency computation purposes. The gradient coming from later stages of the network that is the gradient at output of the network layer, is now also the gradient at the output of the identity, $\bm{g}^{out}$.}
    \label{fig:virtual_id}
\end{figure}

Because of the nature of the identity, this layer does not change the information propagated in either the forward or backward direction.
Conceptually, it is attached to the part of the model that one wishes to inspect as shown in~\cref{fig:virtual_id}.
This layer is never ``physically'' added to the model (i.e., the model is not modified); its ``inclusion'' or ``exclusion'' simply denotes whether we are using input or output activations ($\bm{x}^{in}$ or $\bm{x}^{out}$).
Indeed the backpropagated gradient $\bm{g}^{out}$ at the output of the identity as shown in~\cref{fig:virtual_id} is the gradient that would have been at the output of the layer preceding the identity.
This construction allows to examine activations and gradients at the same location (e.g., $\bm{x}^{out}$ and $\bm{g}^{out}$) as most existing saliency methods do.
We now can use the formulae defined in~\cref{s:extract} when analyzing the gradient of the weights of this identity layer.
For example, if we consider an identity scaling layer, the spatial contribution would be $\bm{g}^{out}_u \odot \bx^{out}_u$. Or, for a $N \times N$ identity convolutional layer, it would be ${\bm{g}^{out}_u}{\bx^{out}_{u, N \times N}}^\top$.

\subsubsection{Phase 2: Aggregate}\label{s:aggregate}

Following the extraction phase~(\cref{s:extract}), we have the local contribution at each spatial location to the gradients of either an existing layer or a virtual identity layer. Each spatial location is associated with a vector or matrix~(\cref{t:layers}). In this section, we describe different aggregation functions to map these vectors or matrices to a single scalar per spatial position. We drop the spatial location $u$ in the notations.

\paragraph{Understanding existing saliency methods.}
One possible aggregation function is the sum of the elements in a vector.
When combined with a virtual scaling identity layer~(\cref{s:virt_ident}), we obtain the linear approximation method~\cite{kindermans2016investigating}: $\sum_k g^{out}_{k} x^{out}_{k}$.
The contributions from the scaling identity encode the result of channels changing (after the gradient is applied) at each location; thus, the $\operatorname{sum}$ aggregation function acts as a voting mechanism.
The resulting saliency map highlights the locations that would be most impacted if following through with the channel updates.

Aggregation functions can also be combined with element-wise filtering functions (e.g., the absolute value function).
Another aggregation function takes the maximum absolute value of the vector: $\operatorname{maxabs}(\bx) = \max_k \lvert x_k \rvert$.
If we combine a virtual bias identity layer in phase 1, which gives  $\bm{g}^{out}$ as the spatial contribution, with the $\operatorname{maxabs}$ function for aggregation, we obtain at each spatial location the gradient~\cite{simonyan14deep} method: $\max_k \lvert g^{out}_{k} \rvert$.

As for CAM~\cite{zhou2016learning} and Grad-CAM~\cite{selvaraju17gradcam}, we cannot directly use the spatial contributions extracted at each location because they spatially average $\bm{g}^{out}$.\footnote{Because CAM  global average pooling + one fully connected layer, $\bm{g}^{out}$ is equal to the fully connected weights.}
However, for architectures that use global average pooling followed after their convolutional layers (e.g., ResNet architectures), $\bm{g}^{out}_u = \bar{\bm{g}}^{out}$.
Then, CAM and Grad-CAM and CAM can be viewed as combining a virtual scaling identity layer from phase 1 with summing and positive filtering (i.e., $\operatorname{filter_{+}}(x) = (x)_{+}$)   functions for aggregation.

\paragraph{NormGrad.} The $\operatorname{sum}$ and $\max$ functions have clear interpretations when using bias or scaling identity layers; however, they cannot be easily transported to convolutional identity layers as interactions between input channels can vary depending on the output channel and are represented as a matrix, not as a vector as are the case for scaling and bias layers.
Thus, we would like to have an aggregation function that could be used to aggregate any type of spatial contribution, regardless of its shape.

Using the $\operatorname{norm}$ function satisfies this criterion, for example the $L^2$ norm when dealing with vectors and the Frobenius norm for matrices.
We note that the matrices obtained at each location for convolutional layers are the outer products of two vectors.
For such matrices, the Frobenius norm is equal to the product of the norms of the two vectors.
For example, for an existing $1 \times 1$ convolutional layer, we consider the saliency map given by $\|{\bm{g}^{out}}\| \|\bx^{in}\|$.
We call this class of saliency methods derived by using $\operatorname{norm}$ as an aggregation function, ``NormGrad''.

Saliency maps from the NormGrad outlined above are not as class selective as other methods because they highlight the spatial locations that contribute the most to gradient of the weights, both positively and negatively.
One way to introduce class selectivity is to use positive filtering before applying the norm.
If we apply $\operatorname{norm}$ and positive filtering aggregation to a scaling identity layer, the resulting saliency map is given by $\| (\bm{g}^{out} \odot \bm{x}^{out})_+ \|$.
Throughout the rest of the paper, we call this variant ``selective NormGrad''.

\begin{table}
\begin{center}
\begin{tabular}{|l|l|l|}
\hline
Phase 1 & Phase 2 & Saliency map \\
\hline
Bias IDL & Max &  $\max_k g^{out}_k$ \\
Scaling IDL & Sum &  $\sum_k g^{out}_k x^{out}_k$ \\
Scaling IDL & F + N &  $\| (g^{out} \odot x^{out})_+ \|$ \\
Conv $1 \times 1$ IDL & Norm & $\|{\bm{g}^{out}}\| \|\bx^{out}\|$ \\
Real Conv $3 \times 3$ & Norm & $\|{\bm{g}^{out}}\| \|{\bx^{in}_{3 \times 3}}\|$ \\
\hline
\end{tabular}
\end{center}
\caption{\textbf{Combining layers and aggregation functions for saliency.}
$g^{out}_k$ and $x^{out}_k$ are tensor slices for channel $k$ and contain only spatial information.
IDL denotes an identity layer. F + N is positive filtering + norm.
From top to bottom, the rows correspond to the following saliency methods: 1., gradient, 2., linear approximation, 3., NormGrad selective, 4., NormGrad, and 5., NormGrad without the virtual identity trick.\label{t:compose}}
\end{table}

%% file: figures/procedure.tex
\begin{algorithm}
\caption{Extract \& Aggregate}\label{f:procedure}
\begin{enumerate}[leftmargin=0pt]
    \item \textbf{Extract.} Compute spatial contributions to the summation of the gradient of the weights.
\Statex \textbullet ~Choose a layer whose parameters are shared spatially (layers from \cref{t:layers}).
\Statex \textbullet ~Alternatively, insert an identity layer~(\cref{s:virt_ident}) at the targeted location.
    \item \textbf{Aggregate.} Transform these spatial contributions into saliency maps using an aggregation function.
\Statex \textbullet ~Norm: NormGrad (ours)
\Statex \textbullet ~Voting/Summing: linear approximation~\cite{kindermans2016investigating}
\Statex \textbullet ~Filtering: GradCAM~\cite{selvaraju17gradcam}, selective NormGrad (ours)
\Statex \textbullet ~Max: Gradient backprogation~\cite{simonyan14deep}
\end{enumerate}
\end{algorithm}

%% file: experiments.tex
\section{Experiments}\label{s:experiments}

In this section we quantitatively and qualitatively evaluate the performance of a large number of backpropagation-based saliency methods. Code for our framework is released at \url{http://github.com/srebuffi/revisiting_saliency}. Additional experiments such as image captioning visualizations are included in the appendix.

\paragraph{Experimental set-up.}
Unless otherwise stated, saliency maps are generated on images from either the PASCAL VOC~\cite{everingham2015pascal} 2007 test set or the ImageNet~\cite{deng2009imagenet} 2012 validation set for either VGG16~\cite{simonyan14deep} or Resnet50~\cite{he2016deep}.
For PASCAL VOC, we use a model pre-trained on ImageNet with fine-tuned fully connected layers on PASCAL VOC.
We use~\cite{fong19}'s TorchRay interpretability package to generate saliency methods for all other saliency methods besides our NormGrad and meta-saliency methods as well as to evaluate saliency maps on~\cite{zhang2016excitation}'s Pointing Game (see ~\cite{zhang2016excitation} for more details).
For all correlation analysis, we compute the Spearman's correlation coefficient~\cite{myers2004spearman} between saliency maps that are upsampled to the input resolution: $224 \times 224$.

\subsection{Justifying the virtual identity trick.}
In order to justify our use of the virtual identity trick, we compare NormGrad saliency maps generated at $1 \times 1$ and $3 \times 3$ convolutional layers both with and without the virtual identity trick (4th and 5th rows respectively in \cref{t:compose}) for VGG16 and ResNet50.
We first computed the correlations between saliency maps generating with and without the virtual identity trick.
We found that the mean correlation across $N = 50k$ ImageNet validation images was over 95\% across all layers we tested.
We also evaluated their performance on the Pointing Game~\cite{zhang2016excitation} and found that the mean absolute difference in pointing game accuracy was $0.53\%\pm 0.62\%$ across all layers we tested (see supp. for more details and full results).
This empirically demonstrates that using the virtual identity trick closely approximates the behavior of calculating the spatial contributions for the original convolutional layers.

\subsection{Combining saliency maps across layers.}

\input{fig-weighting}

Training linear classifiers on top of intermediate representations is a well-known method for evaluating the learned features of a network at different layers~\cite{alain2016understanding}. This suggests that saliency maps, too, may have varying levels of meaningfulness at different layers. 

We explore this question by imposing several weighting methods for combining the layer-wise saliency maps of ResNet50 and VGG16 and measuring the resulting performance on PASCAL VOC Pointing Game on both the ``difficult'' and ``all'' image sets. To determine the weight $\gamma_j$ for a given layer $j$ out of $J$ layers in a network we use:
\begin{enumerate}
\itemsep0em 
\item \textbf{Feature spread}. Given the set of feature activations at layer $j$, $\bx^{i}$ for $i \in M$ input images sampled uniformly across classes, compute the spatial mean $\bar{\bx}^{i} =  \sum_{u\in\Omega} \bx^{i}_{u}$. $\gamma_j = \frac{1}{M} \sum_{i=1}^M | \bar{\bx}^{i} - \bar{\bx}^{\mu}|$ where $\bar{\bx}^{\mu} = \frac{1}{M} \sum_{i=1}^M | \bar{\bx}^{i}|$. 
\item \textbf{Classification accuracy}. Given the set of feature activations at layer $j$, $\bx^{i}$ for $i \in M$ input images sampled uniformly across classes, train a linear layer $\Psi$ using image labels $\by^{i}$. $\gamma_j = \frac{1}{M} \sum_{i=1}^M \delta_{\Psi(\bx^{i}), \by^{i}}$ where $\delta_{\Psi(\bx^{i}), \by^{i}} = 1$ if $\Psi(\bx^{i}) = \by^{i}$, 0 otherwise.
\item \textbf{Linear interpolation}. $\gamma_j = \frac{j}{J}$.
\item \textbf{Uniform}. $\gamma_j = \frac{1}{J}$.
\end{enumerate}
To obtain a combined saliency map $\bbM$ from maps $\bbm_j$ from each layer $j$, the weights are normalised and applied either additively, $\bbM = \sum_{j=0}^J \gamma_j \cdot \bbm_j$, or with a product,
$\bbM = \prod_{j=0}^J \bbm_j^{\gamma_j}$ (see ~\cref{fig:weighting} for visualization).

\input{table1}

\Cref{fig:table1} shows that weighted saliency maps produce the best overall performance in all four key test cases, which is surprising as there were only four weight schemes tested (in addition to best single layer), none of which were explicitly optimised for use with saliency maps. Our results strongly indicate that linear approximation in particular benefits from combining maps from different layers, and linear approximation with layer combination consistently produces the best performance overall and beats far more complex methods at weak localisation using a single forward-backward pass (see supp. for full results).

Note that the feature spread and classification accuracy metrics can both be used as indicators of class sensitivity~(\cref{s:sensitivity}).
This is because if feature activations are uniform for images sampled across classes, it is not possible for them to be sensitive to - or predictive of - class, and the classification accuracy metric is an explicit quantisation of how easily features can be separated into classes.
We observe from the computed weights that both metrics generally increase with layer depth (see supp.). 

\input{fig-ablation}

\paragraph{Explanation of Grad-CAM failure mode.} \Cref{fig:ablation} showed qualitatively that Grad-CAM does not produce meaningful saliency maps at any layer except the last convolutional layer, which is confirmed by Grad-CAM's Pointing Game results at earlier layers. Class sensitivity - as measured by our weighting metrics - increasing with layer depth offers an explanation for this drop in performance. Since Grad-CAM spatially averages the backpropagated gradient before taking a product with activations, each pixel location in the heatmap receives the same gradient vector (across channels) \emph{irrespective} of the image content contained within its receptive field. Thus, if the activation map used in the ensuing product is also not class selective - firing on both dogs and cats for example, \cref{fig:ablation} - the saliency map cannot be. On the other hand, methods that do not spatially average gradients such as NormGrad~(\cref{fig:pointing_partial}) can rely on gradients that are free to vary across the heatmap with underlying class, increasing the class sensitivity of the resulting saliency map.

\subsection{An explicit metric for class sensitivity}\label{s:sensitivity}

\input{fig-class_sensitivity_viz}

\input{fig-class_sensitivity_experiments}

\cite{mahendran16salient} qualitatively shows that early backprop-based methods (e.g., gradient, deconvnet, and guided backprop) are not sensitive to the output class being explained by showing that saliency maps generated w.r.t. different output classes and gradient signals appear visually indistinguishable.
Thus, similar to~\cite{Adebayo2018Sanity}, we introduce a sanity check to measure a saliency method's output class sensitivity.
We compute the correlation between saliency maps w.r.t. to output class predicted with highest confidence (max class) and that predicted with lowest confidence (max and min class respectively) for $N=1000$ ImageNet val. images (1 per class).

We would expect saliency maps w.r.t. the max class to be visually salient while those w.r.t. to the min class to be uninformative (because the min class is not in the image).
Thus, we desire the correlation scores to be close to zero.

\Cref{fig:class_sensitivity_methods} shows results for various saliency methods.
We observe that excitation backprop and guided backprop yield correlation scores close to $1$ for all layers, while contrastive excitation backprop yields scores closest to $0$.
Furthermore, methods using $\operatorname{sum}$ aggregation (e.g., gradient [sum], linear approx, and Grad-CAM) have negative scores (i.e., their max-min-class saliency maps are anti-correlated).
This is because $\operatorname{sum}$ aggregation acts as a voting mechanism; thus, these methods reflect the fact that the network has learned anti-correlated relationships between max and min classes.

\subsection{Meta-saliency analysis}\label{s:order1}
As a general method for improving the sensitivity of saliency heatmaps to the output class used to generate the gradient, we propose to 
perform an inner SGD step before computing the gradients with respect to the loss. This way we can extend any saliency method to second order gradients. This is partly inspired by the inner step used in, for example, few shot learning~\cite{finn2017model} and architecture search~\cite{liu2018darts}. We want to minimize:
\begin{equation}
  L(\theta, x) = \ell(\theta - \epsilon \nabla_\theta \ell(\theta, x), x) .
\end{equation}
We take $\epsilon \ll 1$ to use a Taylor expansion of this loss at $\theta$ and we now have the resulting approximated loss:
\begin{equation}
  L(\theta, x) \approx \ell(\theta, x) - \epsilon \|\nabla_\theta \ell(\theta, x)\|^2 .
\end{equation}
As done in the previous section, we can now take the gradient of the loss with respect to the parameters $\theta$:
\begin{equation}
 \nabla_\theta L(\theta, x) \approx \nabla_{\theta} \ell(\theta, x) - 2\epsilon \nabla^2_\theta \ell(\theta, x) \nabla_{\theta} \ell(\theta, x).
\end{equation}
Using a finite difference scheme of step $h$ as in~\cite{Pearlmutter94fastexact}, we can approximate the hessian-vector product by:
\begin{equation*}
\nabla^2_\theta \ell(\theta, x) \nabla_{\theta} \ell(\theta, x) = \frac{\nabla_{\theta} \ell(\theta, x) - \nabla_{\theta^-} \ell(\theta^-, x)}{h} +O(h).
\end{equation*}
where $\theta^- = \theta - h \nabla_{\theta} \ell(\theta, x)$. We chose on purpose a backward finite difference such that two terms cancel each other when taking $h = 2\epsilon$ and we get:
\begin{equation*}
 \nabla_{\theta}L(\theta, x) \approx \nabla_{\theta'} \ell(\theta', x).
\end{equation*}
where $\theta'= \theta - 2\epsilon \nabla_\theta \ell(\theta, x)$ corresponds to one step of SGD of learning rate $2\epsilon$. We notice that if we take $\epsilon \to 0$, this formula boils back down to the original gradient of the weights without meta step. We further note that this meta saliency approach only requires one more forward-backward pass compared to usual saliency backpropagation methods.

Conversely, if we would like to get an importance map that highlights the degradation of the model's performance, we should add an inner step with gradient ascent within the loss. Hence by minimizing the resulting loss $-\ell(\theta + \epsilon \nabla_\theta \ell(\theta, x), x)$, we get the same formula for the gradients of the weights but with $\theta'= \theta + 2\epsilon \nabla_\theta \ell(\theta, x)$. 


We hypothesize that applying meta-saliency to a saliency method should decrease correlation strength because allowing the network to update one SGD step in the direction of the min class should ``destroy'' the informativeness of the resulting saliency map. We use a learning rate $\epsilon = 0.001$ for the class sensitivity quantitative analysis.
\Cref{fig:class_sensitivity_qual} shows qualitatively that this appears to be the case: without meta-saliency, selective NormGrad and linear approximation yield max (class 2) and min class heatmaps that are highly positively and negatively correlated respectively.
However, when meta-saliency is applied, the min class saliency map appears more random.
\Cref{fig:class_sensitivity_difference} shows results comparing the max-min class correlation scores with and without meta-saliency.
These results demonstrate that meta-saliency decreases max-min class correlation strength for nearly all saliency methods and suggest that meta-saliency can increase the class sensitivity for any saliency method.

\subsection{Model weights sensitivity}

\cite{Adebayo2018Sanity} shows that some saliency methods (e.g., Guided Backprop in particular) are not sensitive to model weights as they are randomized in a cascading fashion from the end to the beginning of the network.
\Cref{fig:sanity} shows qualitatively that, by the late conv layers, saliency maps for linear approximation and selective NormGrad are effectively scrambled (top two rows).
It also highlights that, because meta-saliency increases class selectivity and is allowed to take one SGD in the direction of the target class, it takes relatively longer (i.e., more network depth) to randomize a meta-saliency heatmap (bottom row and see appendix).
\input{fig-sanity}

%% file: fig-weighting.tex
\begin{figure*}[!htb]
    \centering
    \includegraphics[width=0.9\linewidth]{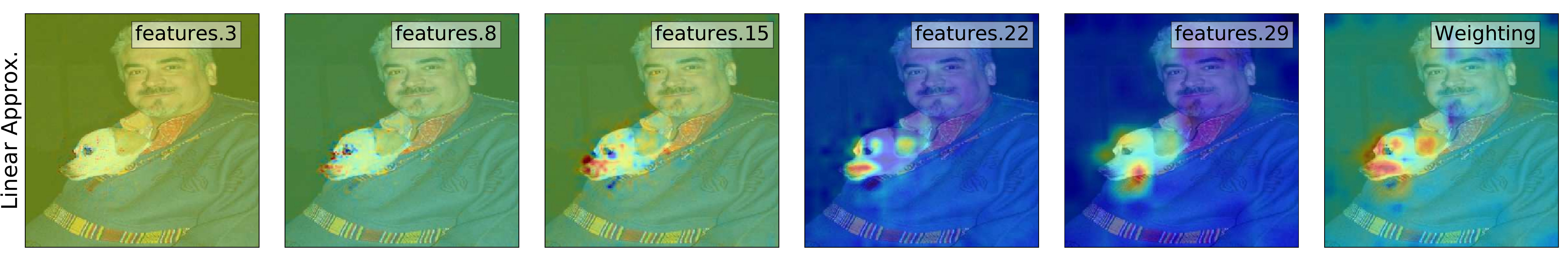}
    \includegraphics[width=0.9\linewidth]{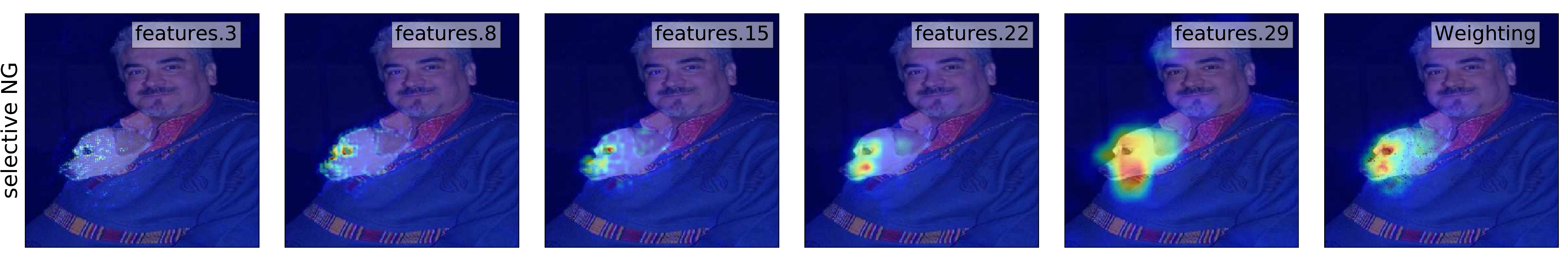}
    \caption{\textbf{Saliency maps at different network depths and as a weighted combination.} Linear approximation vs. selective NormGrad saliency maps of VGG16 on VOC2007. The first 5 images of each row correspond to different depths whereas the last one is a weighted product combination (using classification accuracy weights) of the first saliency maps. We observe that the weighted version produces more fined grained maps for both methods.} 
    \label{fig:weighting}
\end{figure*}

%% file: table1.tex
\begin{table}
\centering
\setlength\tabcolsep{3pt}
\begin{tabular}{|l|c|c|c|c|c|c|c|c|}
\hline
 & \multicolumn{4}{|c|}{All} & \multicolumn{4}{|c|}{Difficult} \\
 \hline
 & \multicolumn{2}{|c|}{Resnet50} & \multicolumn{2}{|c|}{VGG16} & \multicolumn{2}{|c|}{Resnet50} & \multicolumn{2}{|c|}{VGG16} \\ 
\hline
 & b.s. & b.w. & b.s. & b.w. & b.s. & b.w. & b.s. & b.w.  \\
\hline
CEB & \bfseries 90.7 &           88.6 &           82.1 &           78.2 &            82.2 &           82.2 &            67.0 &           65.2 \\     
EB  &           84.5 &           83.1 &           77.5 &           75.7 &            71.5 &           71.3 &            57.8 &           56.1 \\   
GC  &           90.3 &           90.5 & \bfseries 86.6 &           80.6 &  \bfseries 82.3 &           82.6 &            74.0 &           67.8 \\   
Gd  &           83.9 &           83.3 & \bfseries 86.6 &           82.7 &            70.3 &           69.4 &            66.4 &           67.4 \\   
Gds &           80.0 &           77.4 &           76.8 &           77.2 &            62.9 &           59.5 &            57.9 &           59.4 \\   
Gui &           82.3 &           81.0 &           75.8 &           74.4 &            67.9 &           63.4 &            53.0 &           51.6 \\  
LA  &           90.2 & \bfseries 91.2 &           86.4 & \bfseries 86.9 &            81.9 & \bfseries 83.8 &  \bfseries 74.5 & \bfseries 77.4 \\   
NG  &           84.6 &           83.5 &           81.9 &           81.8 &            72.2 &           70.2 &            64.8 &           64.6 \\   
sNG &           87.4 &           88.7 &           86.0 &           86.8 &            77.0 &           79.1 &            72.6 &           74.5 \\   
\hline
\end{tabular}
\caption{\textbf{Pointing game results on VOC07.} b.s. and b.w. stand for best single layer and best weighted combination. (C)EB: (Contrastive) Excitation Backprop, GC: GradCAM, Gd(s): Gradient (sum), Gui: guided backprop, LA: linear approximation, (s)NG: (selective) NormGrad.}
\label{fig:table1}
\end{table}

\begin{figure}[!htb]
    \centering
    \includegraphics[width=0.9\linewidth]{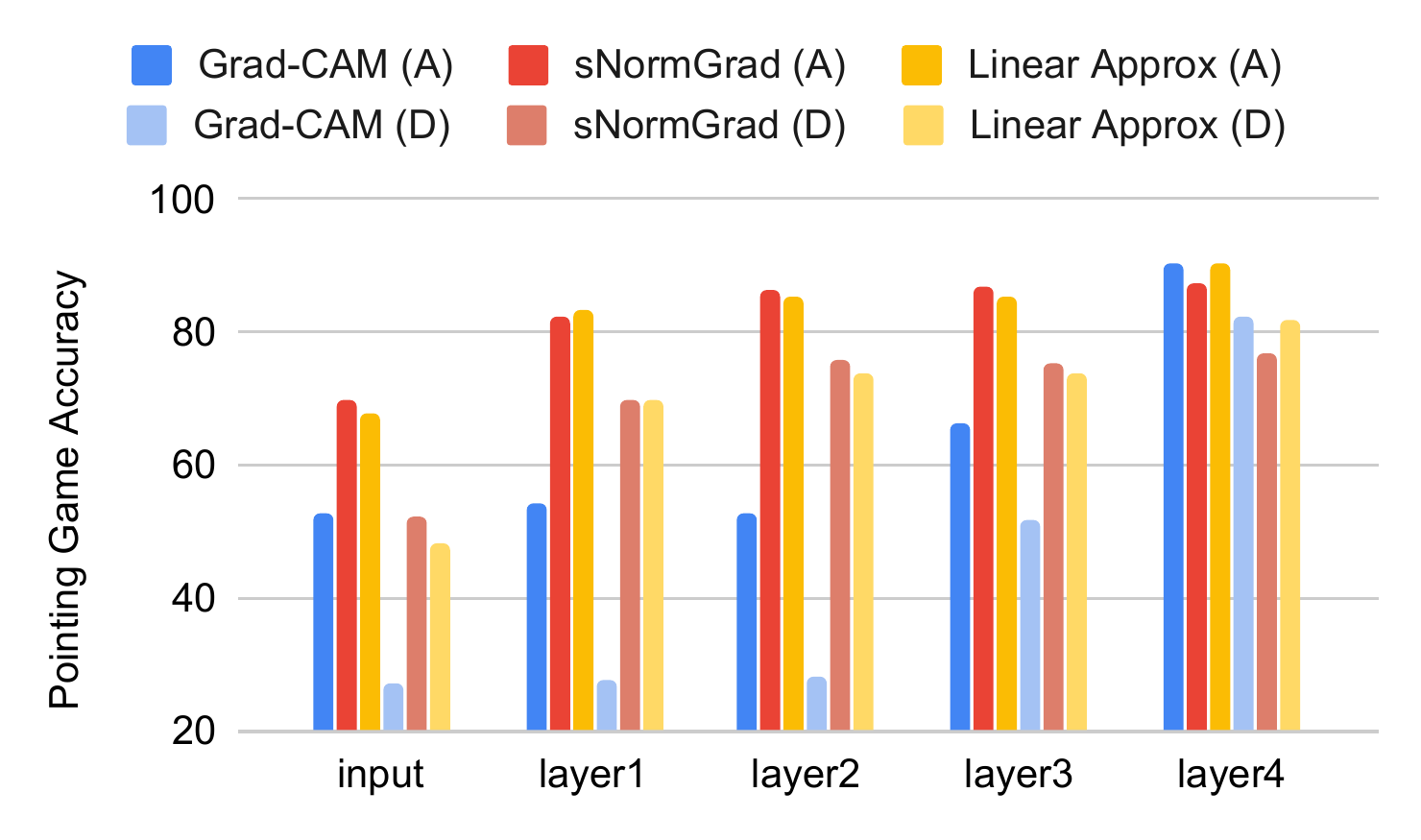}
    \caption{\textbf{Select Pointing Game results.} Results for ResNet50 on VOC07 at different network depths (A: all images; D: difficult subset). Grad-CAM performs worse at every layer except the last conv layer and lower than pointing at the center (all: 69.6\%; diff: 42.4\%) at most layers.}
    \label{fig:pointing_partial}
\end{figure}

%% file: fig-ablation.tex
\begin{figure}[!htb]
    \centering
    \includegraphics[width=\linewidth]{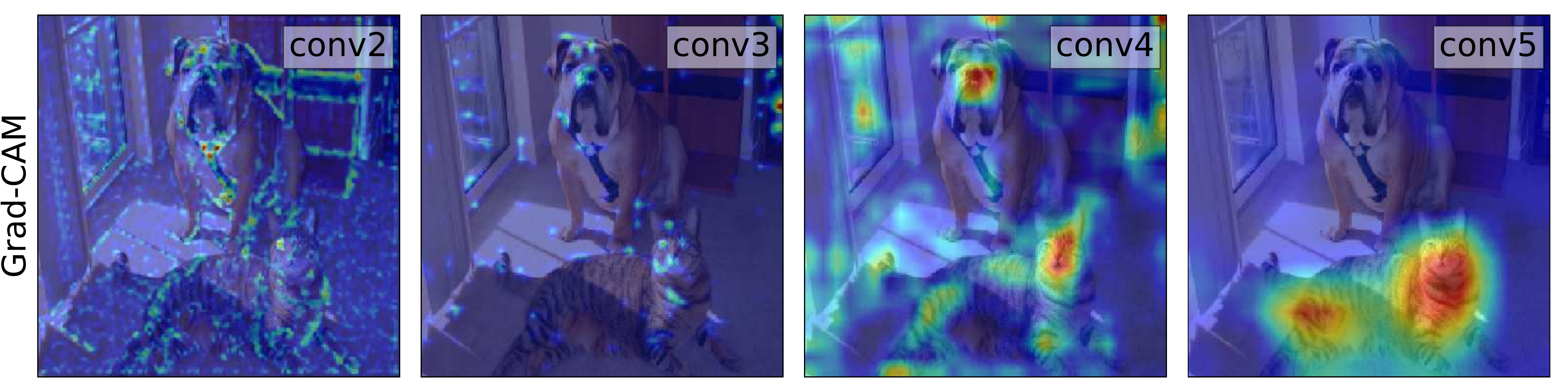}
    \caption{\textbf{Grad-CAM failure mode.} Grad-CAM saliency maps w.r.t. ``tiger cat'' at different depths of VGG16. Grad-CAM only works at the last conv layer (rightmost col).} 
    \label{fig:ablation}
\end{figure}

%% file: fig-class_sensitivity_viz.tex
\begin{figure}[!htb]
    \centering
    \includegraphics[width=0.9\linewidth]{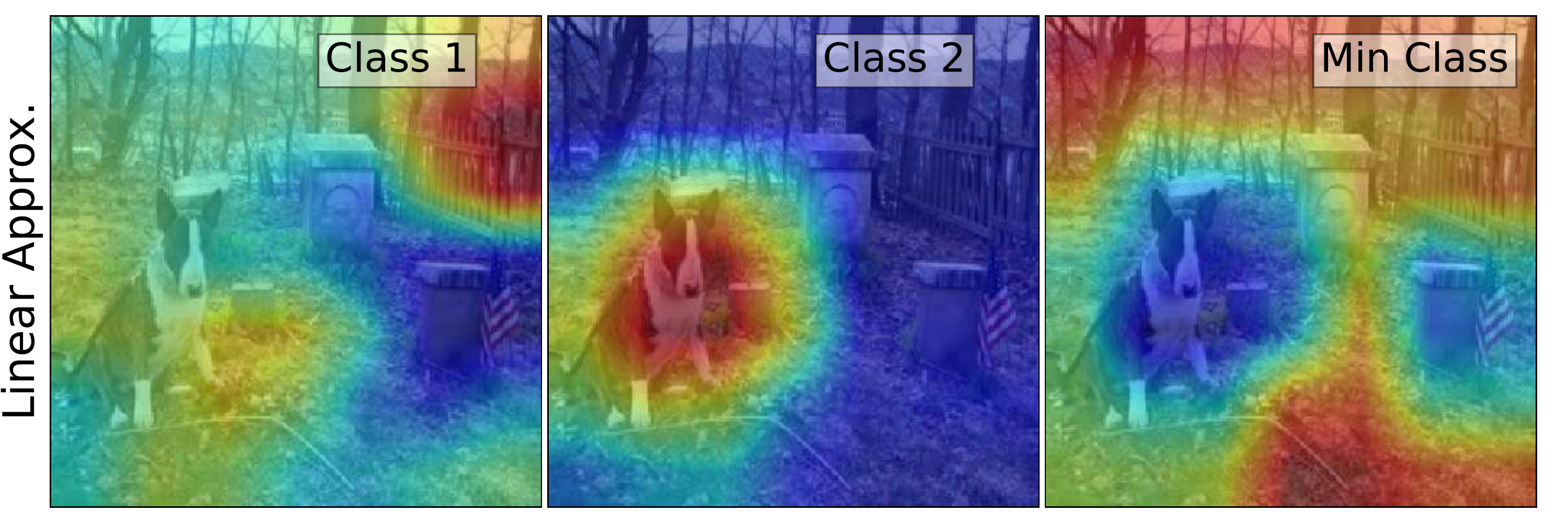}
    \includegraphics[width=0.9\linewidth]{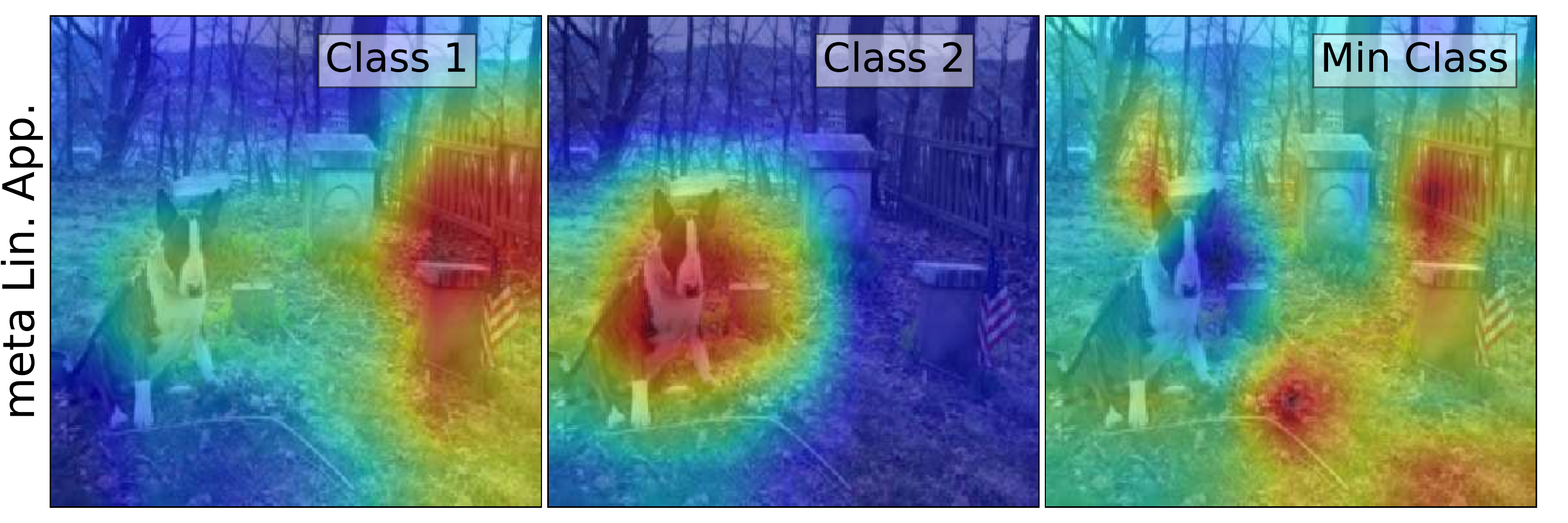}
    \includegraphics[width=0.9\linewidth]{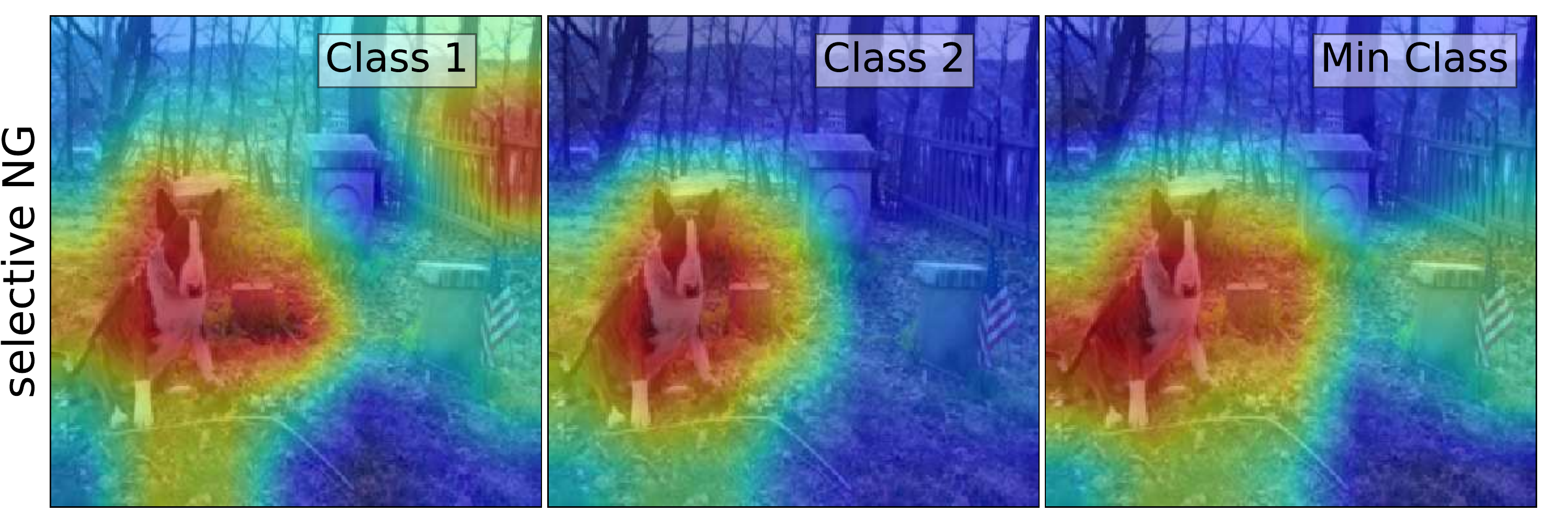}
    \includegraphics[width=0.9\linewidth]{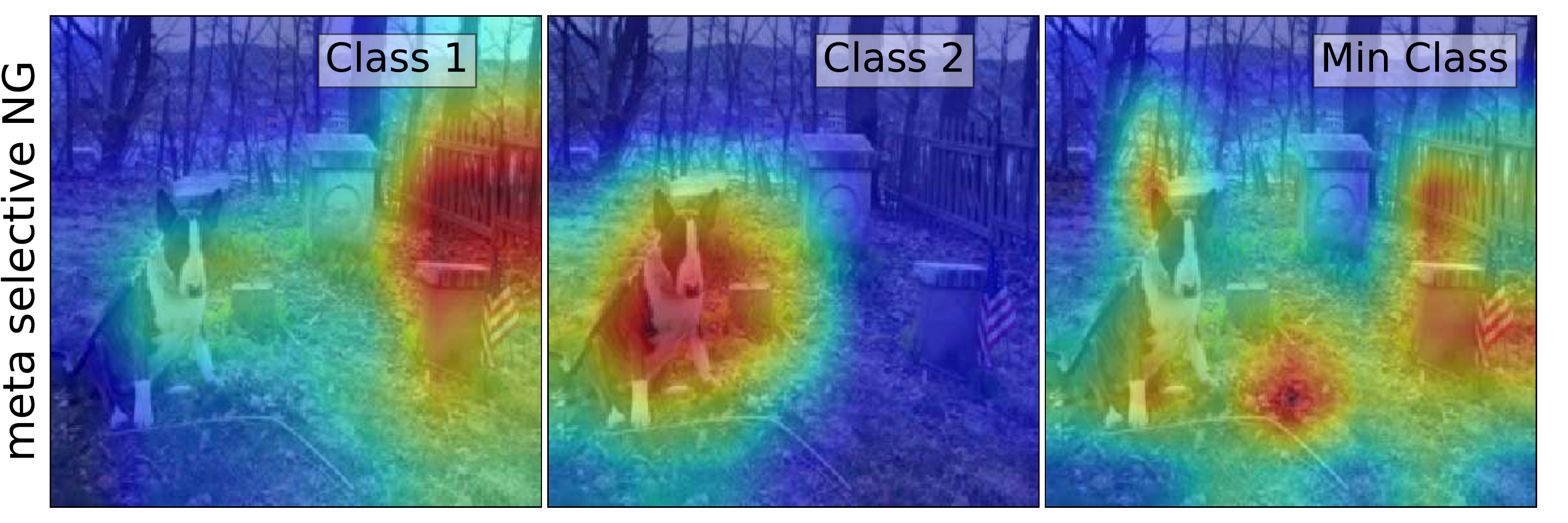}
    \caption{\textbf{Class sensitivity with and without meta-saliency.}
    Min class saliency maps that use meta-saliency (row 2 and 4, right col) are less informative than those that don't use meta-saliency (rows 1 and 3, right col).
    Class 1 is the ground truth class (fence), class 2 is the maximally predicted class (Cardigan Welsh corgi), min class is the minimally predicted class (black widow spider).
    } 
    \label{fig:class_sensitivity_qual}
\end{figure}

%% file: fig-class_sensitivity_experiments.tex
\begin{figure}[!htb]
    \centering
    \includegraphics[width=0.9\linewidth]{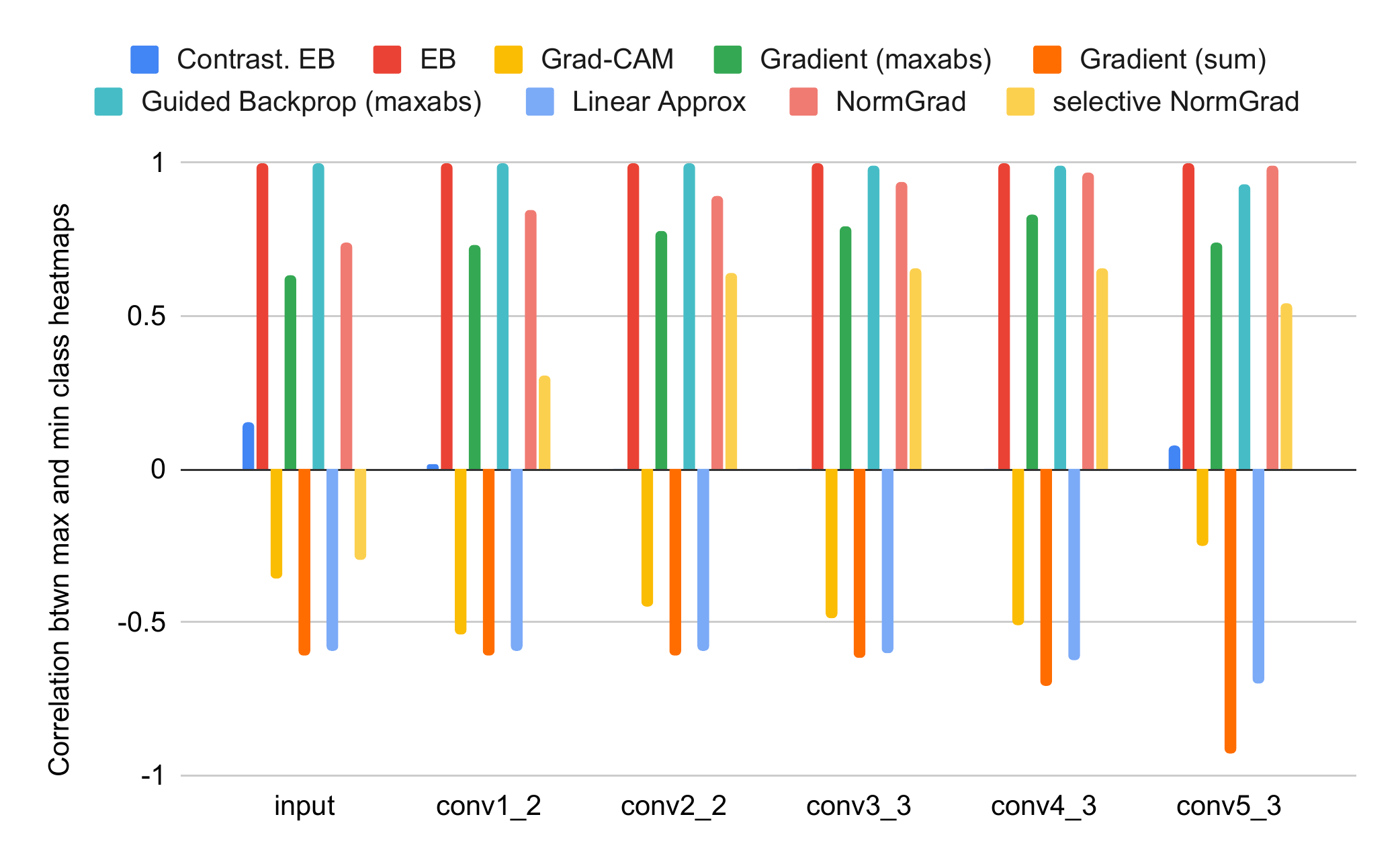}
    \caption{\textbf{Class sensitivity of saliency methods.}
    This plot shows the correlation between VGG16 saliency maps computed w.r.t. to the maximally and minimally predicted class (closer to zero is better).}
    \label{fig:class_sensitivity_methods}
\end{figure}

\begin{figure}[!htb]
    \centering
    \includegraphics[width=0.9\linewidth]{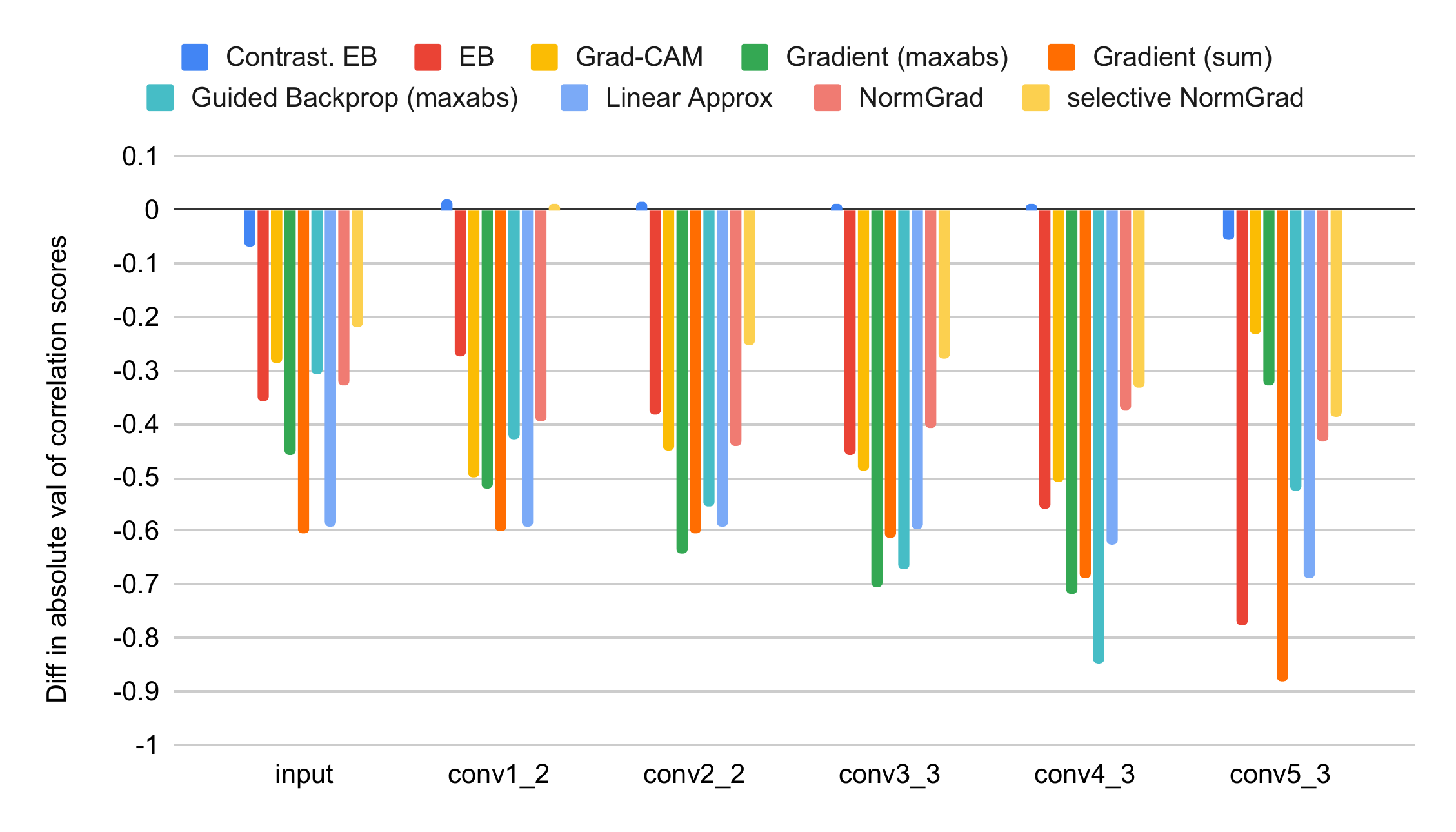}
    \caption{\textbf{Meta-saliency improves class sensitivity for all saliency methods.} Using meta-saliency yields weaker correlations between the saliency maps w.r.t. the maximally and minimally predicted output class compared to not using meta-saliency (lower is better).}
    \label{fig:class_sensitivity_difference}
\end{figure}

%% file: fig-sanity.tex
\begin{figure}[!htb]
    \centering
    \includegraphics[width=\linewidth]{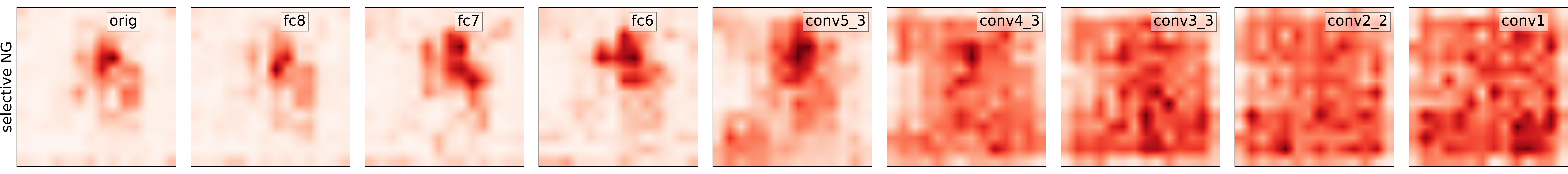}
    \includegraphics[width=\linewidth]{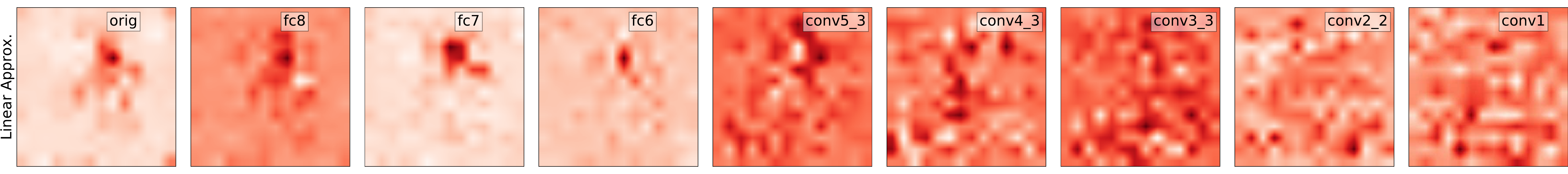}
    \includegraphics[width=\linewidth]{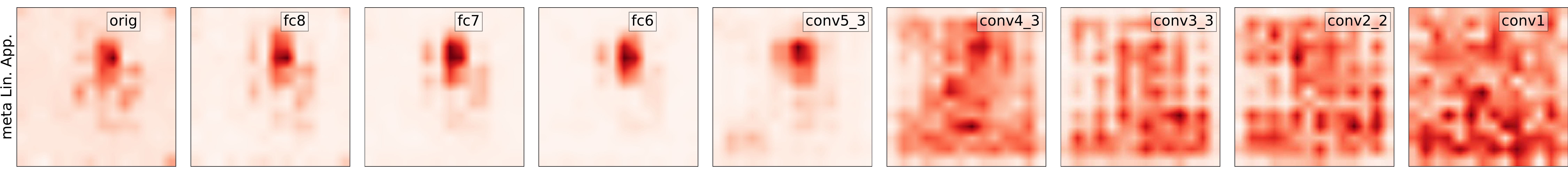}
    \caption{\textbf{Model weights sensitivity.} Sanity check by randomizing VGG16 model weights in a cascading fashion for the ``Irish terrier'' image from~\cite{Adebayo2018Sanity}. Top row: selective NormGrad, middle row: linear approximation, bottom row: linear approximation with meta-saliency (lower correlation with orig heatmap [leftmost col] is better). All methods look random after conv4\_3.
    By comparing the last two rows at conv5\_3, we see the that meta-saliency enforces more class sensitivity than the non-meta variant.
    } 
    \label{fig:sanity}
\end{figure}

%% file: conclusions.tex
\section{Conclusions}\label{s:conclusions}
We introduced a principled framework based on the contribution of each spatial location to the weights' gradient. This framework unifies several existing backpropagation-based methods and allowed us to systematically explore the space of possible saliency methods. We use it for example to formulate NormGrad, a novel saliency method. We also studied how to combine saliency maps from different layers, discovering that it can consistently improve weak localization performance and produce high resolution maps. Finally, we introduced a class-sensitivity metric and proposed meta-saliency, a novel paradigm applicable to any existing method to improve sensitivity to the target class. 

\section{Acknowledgments}
This work is supported by Mathworks/DTA, Open Philanthropy, EPSRC AIMS CDT and ERC 638009-IDIU.

%% file: supps.tex
\clearpage

\twocolumn[
\begin{centering}
\section*{There and Back Again: Revisiting Backpropagation Saliency Methods\\ Supplementary Material}\label{s:suppl}
\end{centering}
\vspace*{10mm}
]

\section{Pointing Game details}
We use ~\cite{zhang2016excitation}'s Pointing Game for quantitative evaluation of saliency methods.
Saliency maps are computed with respect to every object class present in each image.
If the maximally salient point for each class lands on the ground truth annotation for that object (within a threshold of 15 pixels), then a ``hit'' is recorded; otherwise, a ``miss'' is recorded.
Pointing game accuracy is computed as the mean over per-class accuracies given by the following: $\frac{\lvert\text{hits}\rvert}{\lvert\text{hits}+\text{misses}\rvert}$.

We evaluate VGG16 and ResNet50 networks that have been trained on ImageNet and fine-tuned on PASCAL VOC and COCO.
We evaluate on the PASCAL VOC 2007 test split ($N = 4952$ images) and COCO 2014 val split ($N \approx 50k$).
We also show performance on the difficult subsets of the data provided by~\cite{zhang2016excitation}; these are images for which the total area of the annotations (bounding boxes for PASCAL VOC and segmentation masks for COCO) for the given object class is less than $25\%$ of the image size and for which there is at least one other object class present.
We use~\cite{fong19}'s TorchRay library for evaluation; see~\cite{zhang2016excitation} for more details.


\section{ Virtual identity trick correlations}

We computed the correlation between saliency maps generated both with and without the virtual identity trick (paper section 4.1). The high correlation shown in~\Cref{fig:table_id_noid}, as well as the minimal difference in pointing game performance ($0.53\%\pm 0.62\%$) demonstrates that the identity trick closely approximates the behaviour of calculating the spatial contributions for the original convolutional layers.

\begin{table}[h!]
\centering
\begin{tabular}{|l|l|c|}
\hline
Architecture & Layer & Correlation \\
\hline
ResNet50 & layer1.0.conv1 & 99.48 $\pm$ 0.55 \\
ResNet50 & layer2.0.conv1 & 99.32 $\pm$ 0.35 \\
ResNet50 & layer3.0.conv1 & 99.16 $\pm$ 0.42 \\
ResNet50 & layer4.0.conv1 & 98.35 $\pm$ 1.05 \\
VGG16 & features.2 & 97.77        $\pm$ 1.34 \\
VGG16 & features.7 & 99.15        $\pm$ 0.42 \\
VGG16 & features.14 & 98.99       $\pm$ 0.57 \\
VGG16 & features.21 & 97.79       $\pm$ 1.42 \\
VGG16 & features.28 & 94.33       $\pm$ 3.09 \\
\hline
\end{tabular}
\caption{\textbf{Correlations between saliency maps generated with and without the virtual  identity  trick.} For both VGG16 and ResNet50, the high values of correlation across the network (min of 94.33) justify the use of the identity trick.}
\label{fig:table_id_noid}
\end{table}

\FloatBarrier

\section{ Performance of combining saliency maps}

\begin{figure}[!h]
    \centering
    \includegraphics[width=\linewidth]{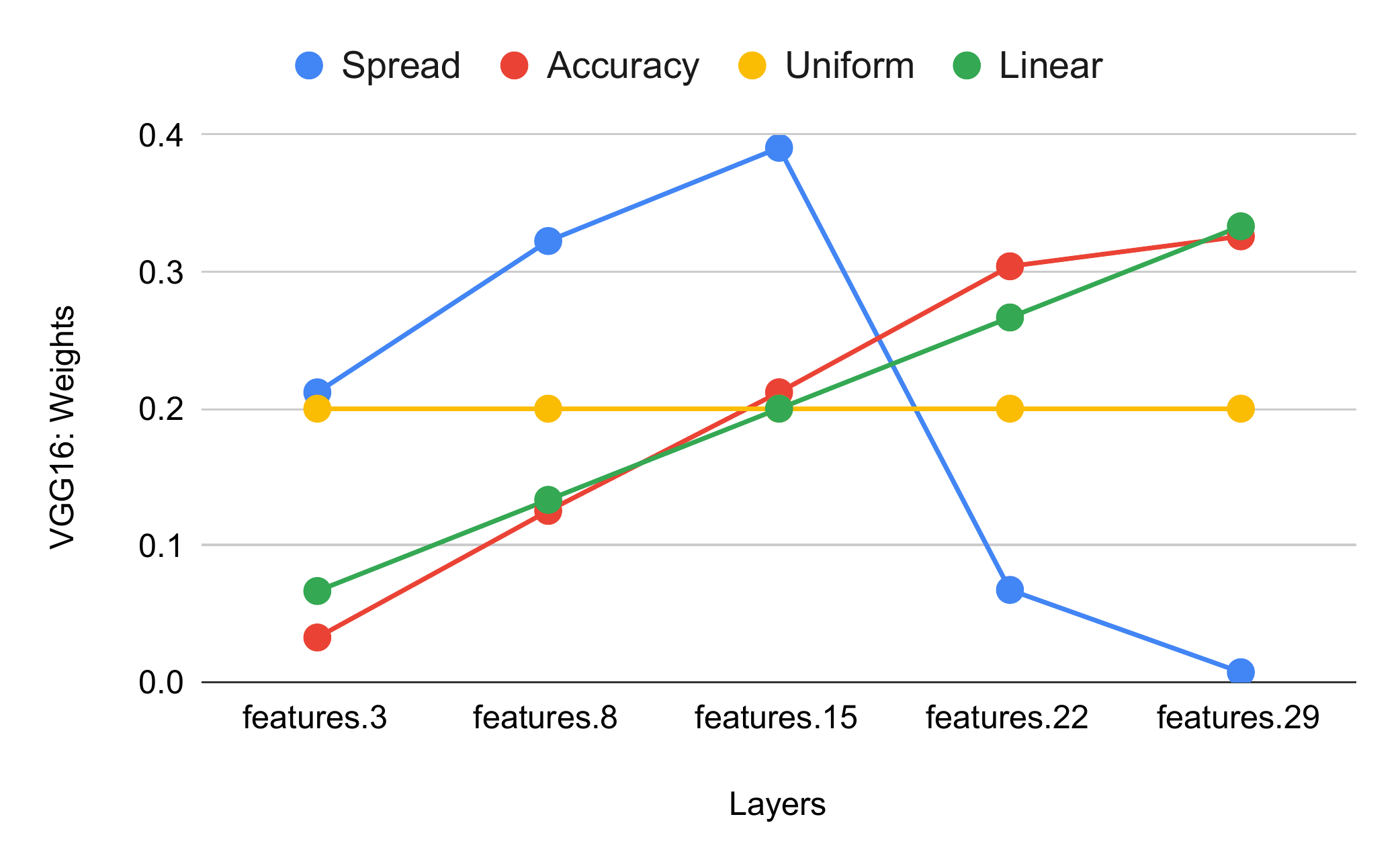}
    \includegraphics[width=\linewidth]{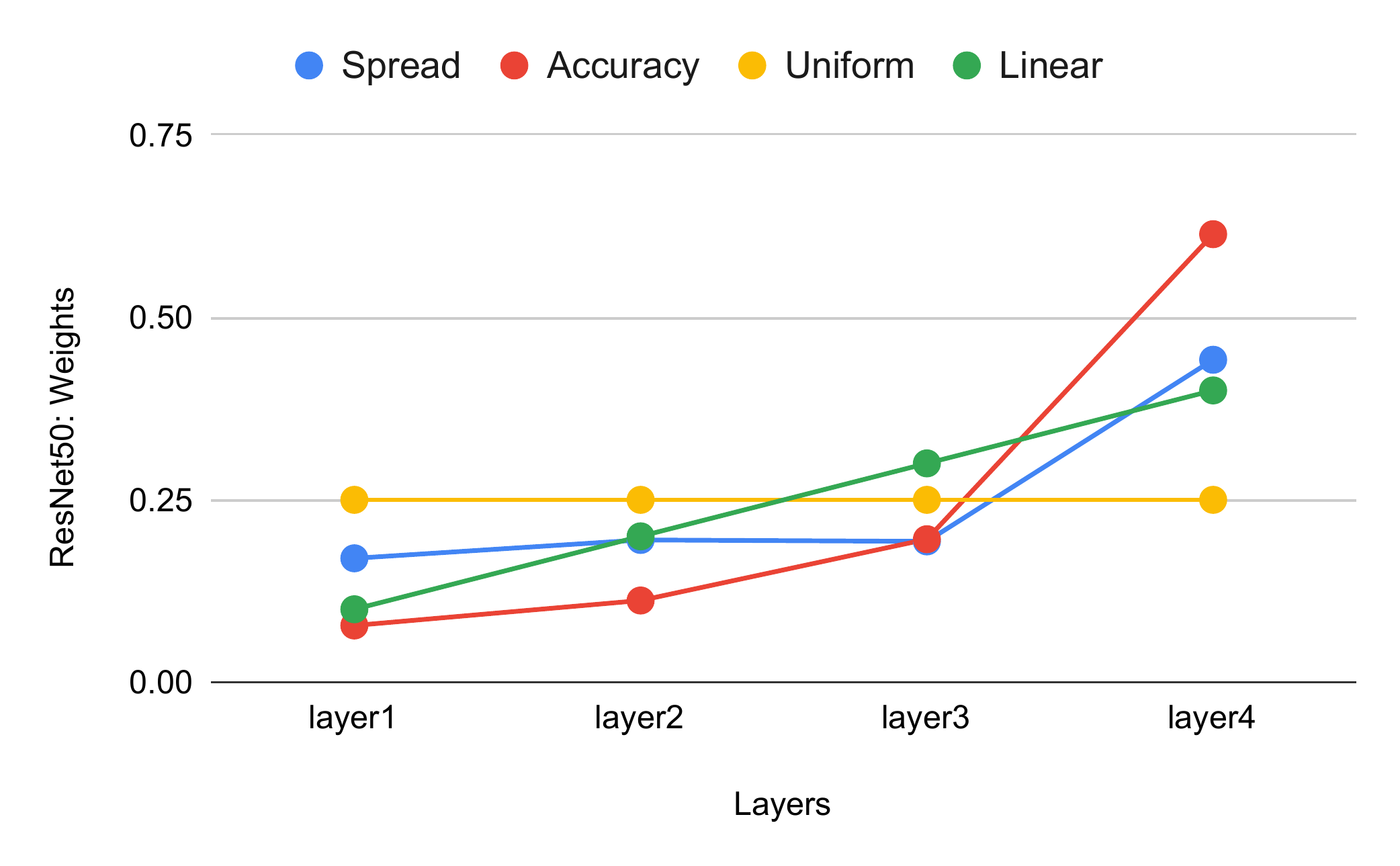}
    \caption{\textbf{Weights for the different weighting methods.} VGG16 on top and ResNet50 on the second row.} 
    \label{fig:weights_layers}
\end{figure}

As noted in paper section 4.2, feature spread and classification accuracy are both interpretable as measures of feature sensitivity with respect to the class of the input image.
~\Cref{fig:weights_layers} shows that both increase with network depth, with the exception of feature spread for VGG16, which decreases in the last two layers despite simultaneously increasing classification accuracy. 
We note that classification accuracy is a more reliable metric than feature spread as it directly codes for the separability of features with respect to class, whereas feature spread is susceptible to non-material differences in the absolute scale of activation values across layers. 
In the case of VGG16, this means the features in the last two layers differ less across images (and hence, classes) in absolute terms, but are nonetheless highly discriminative of class.

The performance gains of our weighting schemes over using the best individual layer are given in~\cref{fig:best_combi}.
The best individual layers for computing maps are given in~\cref{fig:best_layer}, which is notable as in no case does the practice of computing saliency at the earliest layer produce the best performance.

\begin{table}
\centering
\begin{tabular}{|l|c|c|c|c|}
\hline
 & \multicolumn{2}{|c|}{ResNet50} & \multicolumn{2}{|c|}{VGG16} \\
 \hline
 & All & Difficult & All & Difficult\\ 
\hline
CEB & layer3 & layer4 & feat.29 & feat.29 \\     
EB  & layer4 & layer4 & feat.29 & feat.29 \\   
GC  & layer4 & layer4 & feat.29 & feat.29 \\   
Gd  & layer2 & layer2 & feat.29 & feat.22 \\   
Gds & layer2 & layer2 & feat.8 & feat.8 \\   
Gui & layer3 & layer3 & input & input \\  
LA  & layer4 & layer4 & feat.29 & feat.29 \\   
NG  & layer3 & layer3 & feat.22 & feat.22 \\   
sNG & layer4 & layer4 & feat.29 & feat.29 \\   
\hline
\end{tabular}
\caption{\textbf{Best individual layer for the Pointing game on VOC07.} (C)EB: (Contrastive) Excitation Backprop, GC: GradCAM, Gd(s): Gradient (sum), Gui: guided backprop, LA: linear approximation, (s)NG: (selective) NormGrad.}
\label{fig:best_layer}
\end{table}

\begin{table}
\centering
\setlength\tabcolsep{3pt}
\begin{tabular}{|l|l|l|c|c|r|r|}
\hline
 & & & \multicolumn{2}{|c|}{ResNet50} & \multicolumn{2}{|c|}{VGG16} \\
 \hline
 &  & & All & Difficult & All & Difficult\\ 
\hline
LA & $+$ & accuracy & 0.57 & 1.47 & 0.22 & 2.09 \\
LA & $+$ & spread & 0.81 & 1.66 & -3.01 & -7.18 \\
LA & $+$ & linear & 0.68 & 0.74 & 0.36 & 2.22 \\
LA & $+$ & uniform & 0.58 & 1.29 & -0.61 & -0.20 \\
LA & $\times$ & accuracy & 0.87 & \bfseries{1.84} & 0.42 & \bfseries{2.91} \\
LA & $\times$ & spread & \bfseries{1.01} & 1.78 & -2.92 & -6.80 \\
LA & $\times$ & linear & 0.52 & 0.72 & \bfseries{0.49} & 2.87 \\
LA & $\times$ & uniform & 0.32 & 0.83 & -0.25 & 0.52 \\
sNG & $+$ & accuracy & 1.08 & 1.37 & 0.48 & 0.85 \\
sNG & $+$ & spread & 0.94 & 1.40 & -3.66 & -6.43 \\
sNG & $+$ & linear & 0.94 & 1.67 & 0.62 & 0.85 \\
sNG & $+$ & uniform & 0.25 & 0.54 & -0.93 & -2.55 \\
sNG & $\times$ & accuracy & \bfseries{1.26} & 1.65 & \bfseries{0.83} & \bfseries{1.88} \\
sNG & $\times$ & spread & 1.12 & \bfseries{2.09} & -3.13 & -6.18 \\
sNG & $\times$ & linear & 0.63 & 0.95 & 0.72 & 0.62 \\
sNG & $\times$ & uniform & 0.89 & 1.79 & -0.27 & -1.76 \\
\hline
\end{tabular}
\caption{\textbf{Score gains compared to best individual layer performance for weighting methods with Linear Approximation (LA) and selective NormGrad (sNG) on the Pointing game on VOC07.} Paper showed that LA and sNG benefit from weighting methods. Here we can observe that using the weights as exponents in a product (lines with $\times$) is the most effective solution for both LA and sNG on ResNet50 and VGG16. Both the weightings using the features spread and the layer accuracy perform the best on ResNet50 but only the layer accuracy perform consistently across datasets and saliency methods.}
\label{fig:best_combi}
\end{table}

\section{Meta-saliency analysis}

We hypothesized that meta vs. non-meta saliency maps should be less correlated for validation images than for training images. This is because the inner gradient step of meta-saliency should not be as impactful for a seen training image as for an unseen validation image. 

\begin{figure}[!htb]
    \centering
    \includegraphics[width=\linewidth]{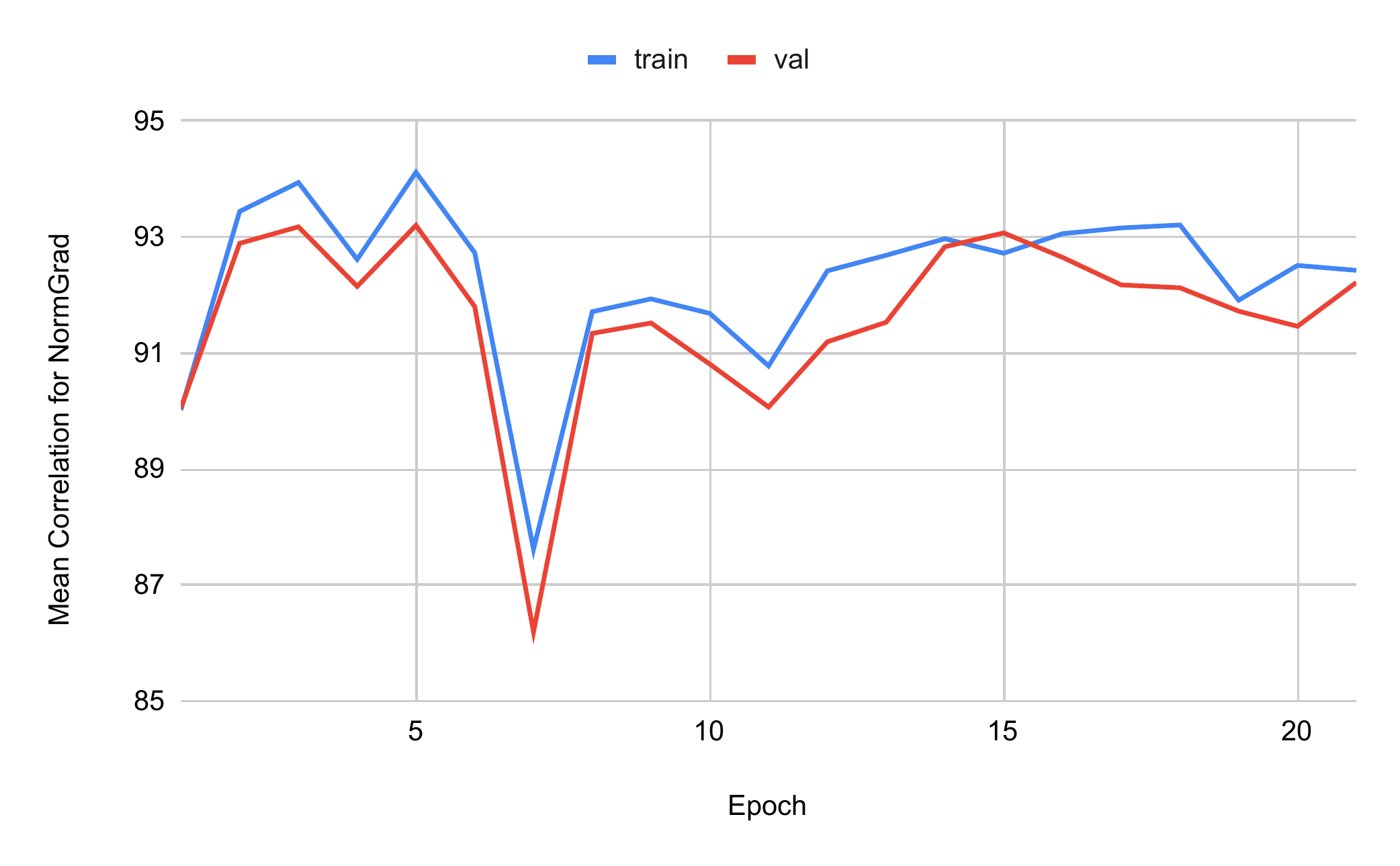}
    \includegraphics[width=\linewidth]{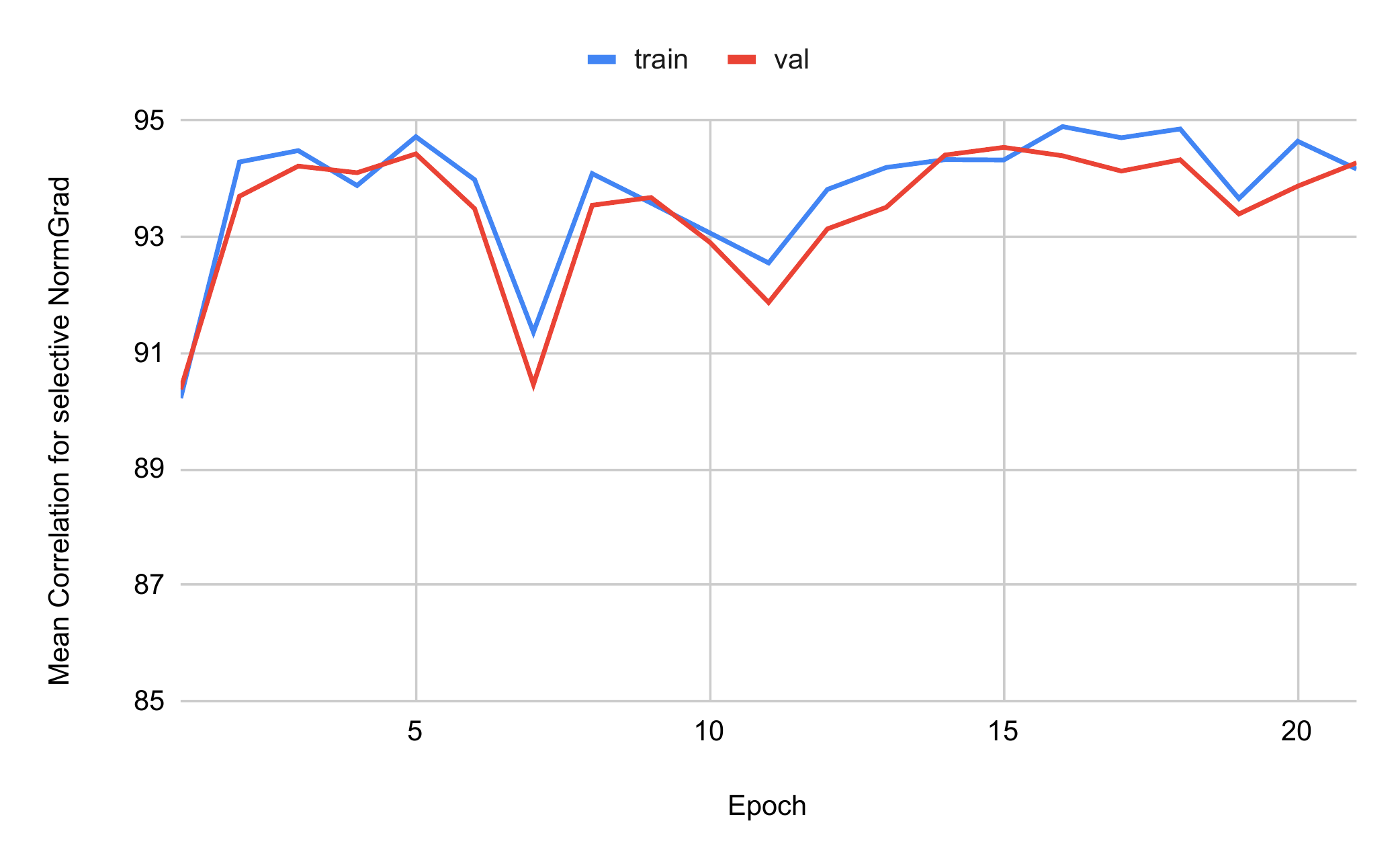}
    \caption{\textbf{Correlation between meta and non-meta saliency maps over time for train and val splits.} For both NormGrad (top) and selective NormGrad (bottom), the correlation is lower on the validation split.}  
    \label{fig:corr_train_val_time}
\end{figure}

For both NormGrad and selective NormGrad, we evaluated over time the average correlation between the importance maps with and without meta-saliency on the training and testing sets respectively.~\Cref{fig:corr_train_val_time} demonstrates that the correlation scores are indeed decreased for validation images.

\begin{figure}[!htb]
    \centering
    \includegraphics[width=\linewidth]{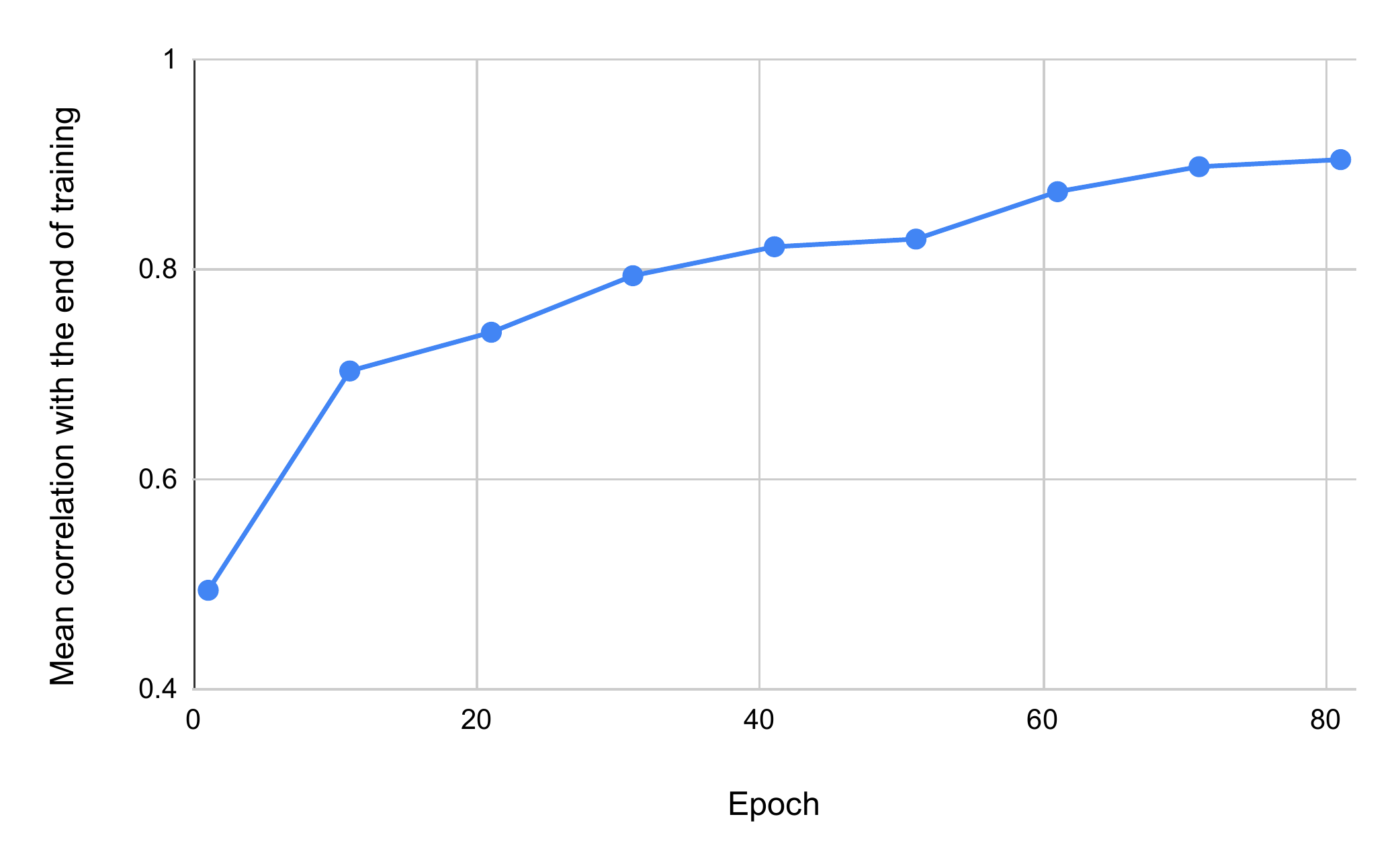}
    \caption{\textbf{Mean correlation on the val split between the meta-saliency maps of selective NormGrad at epoch $t$ and the end of training.} The saliency maps stabilize at the end of the training.} 
    \label{fig:corr_end}
\end{figure}

\begin{figure}[!htb]
    \centering
    \includegraphics[width=\linewidth]{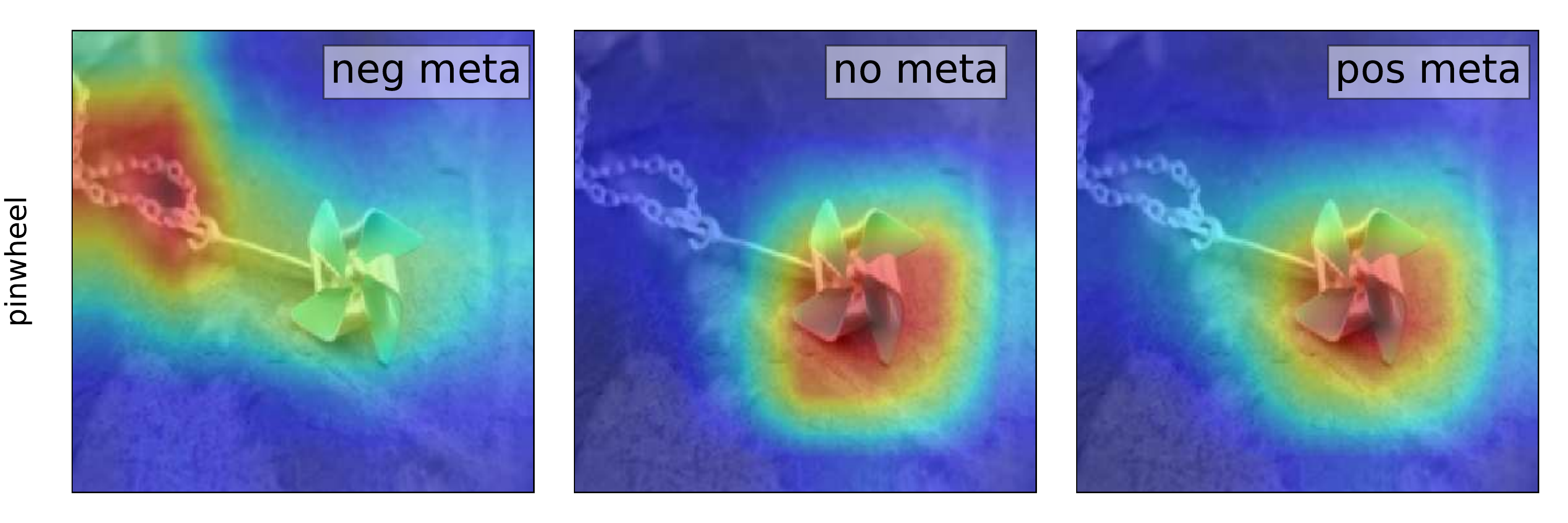}
    \includegraphics[width=\linewidth]{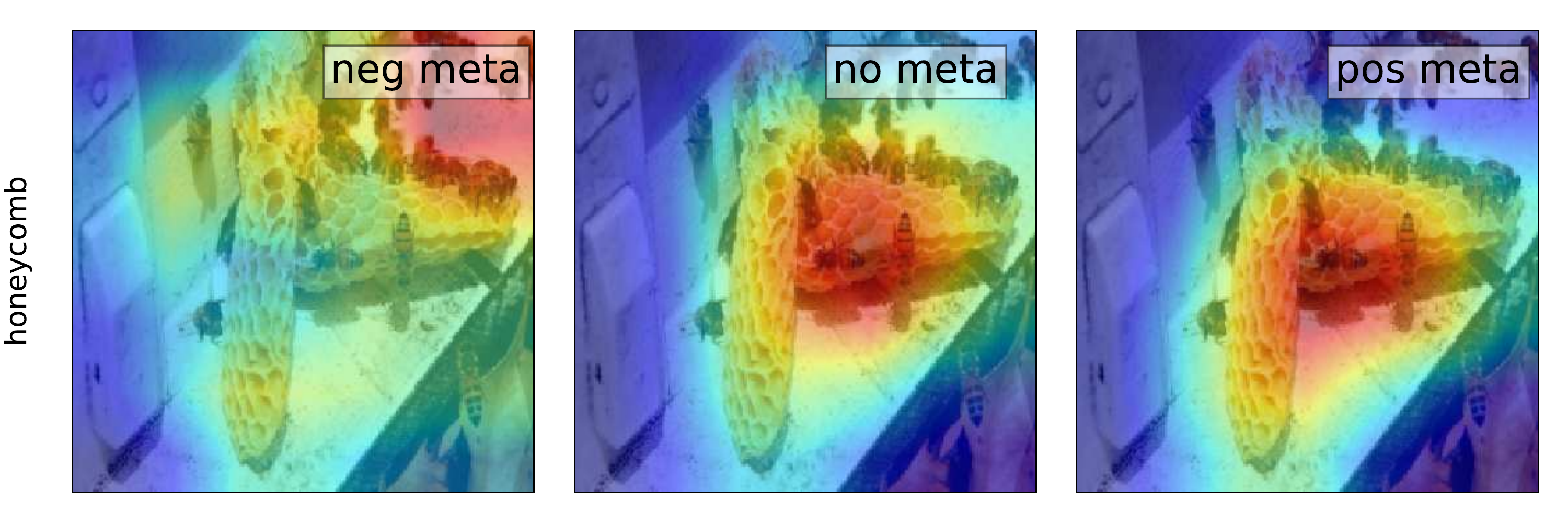}
    \includegraphics[width=\linewidth]{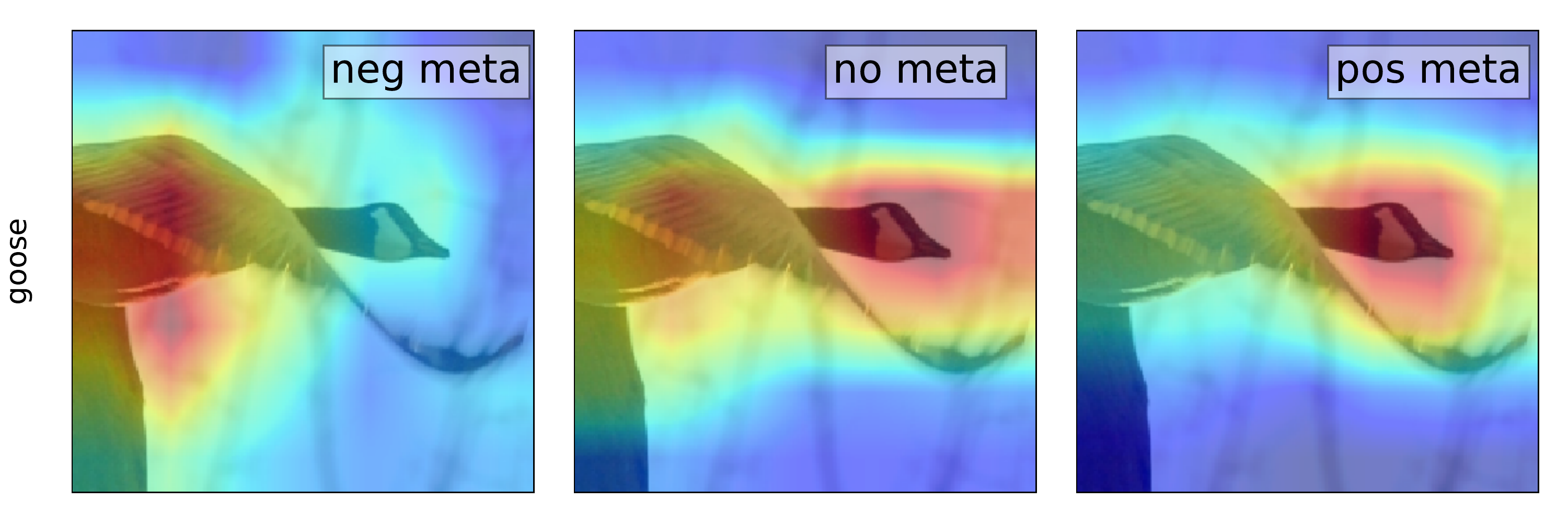}
    \includegraphics[width=\linewidth]{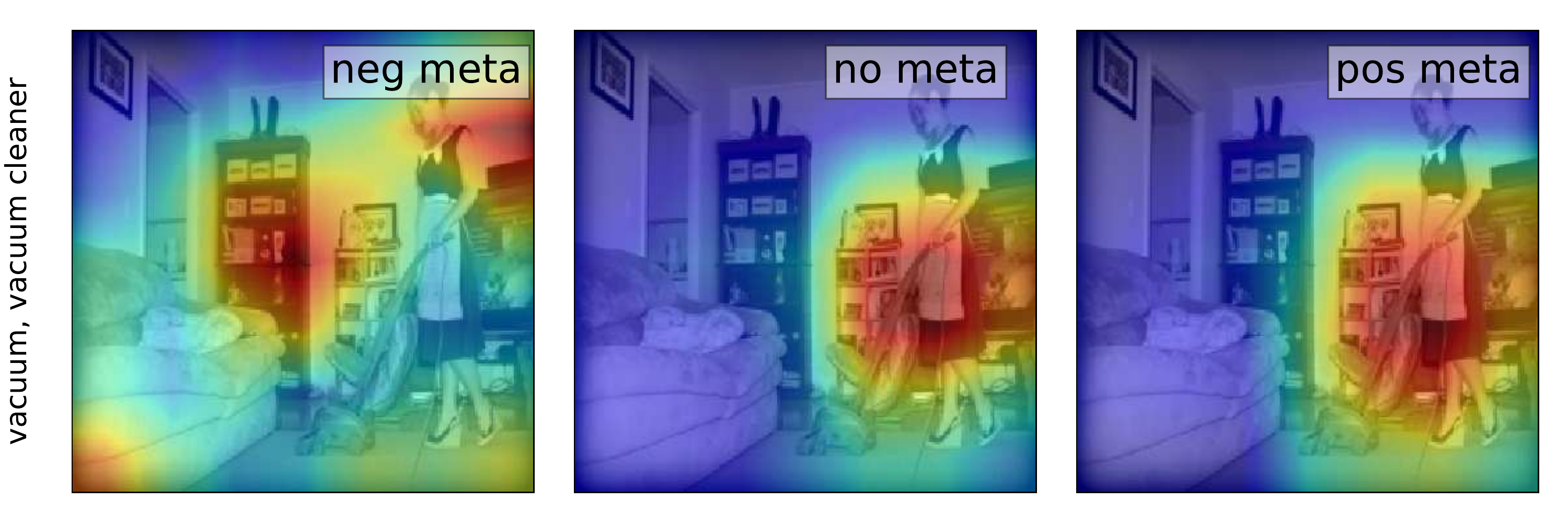}
    \includegraphics[width=\linewidth]{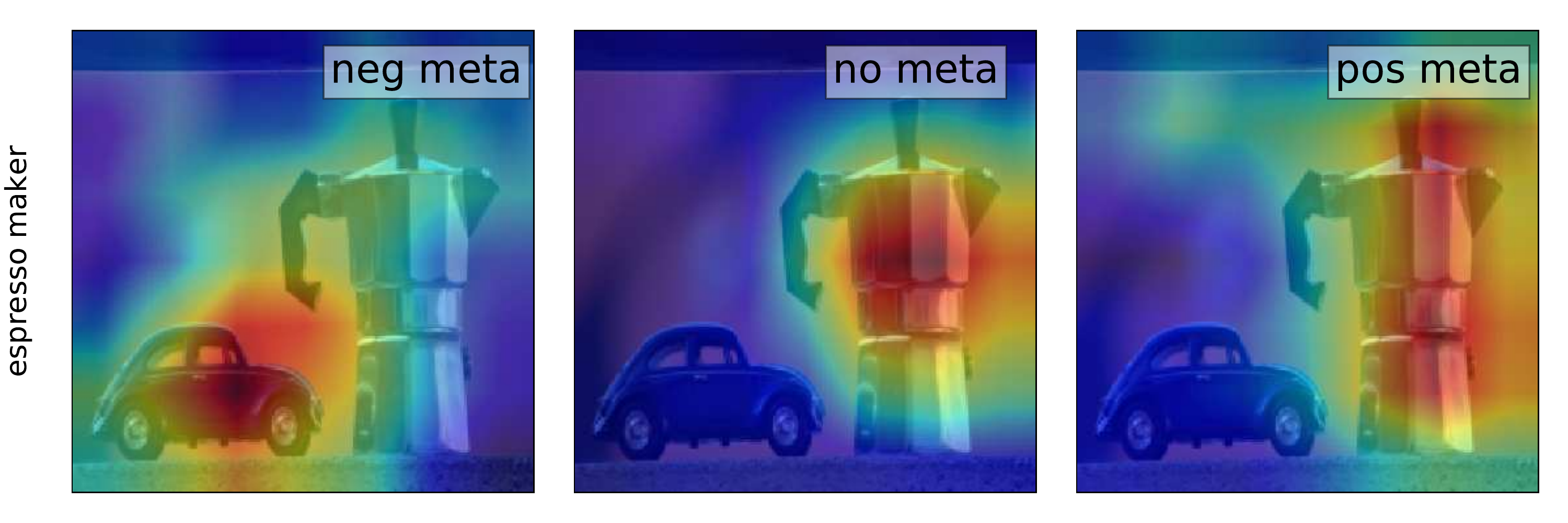}
    \includegraphics[width=\linewidth]{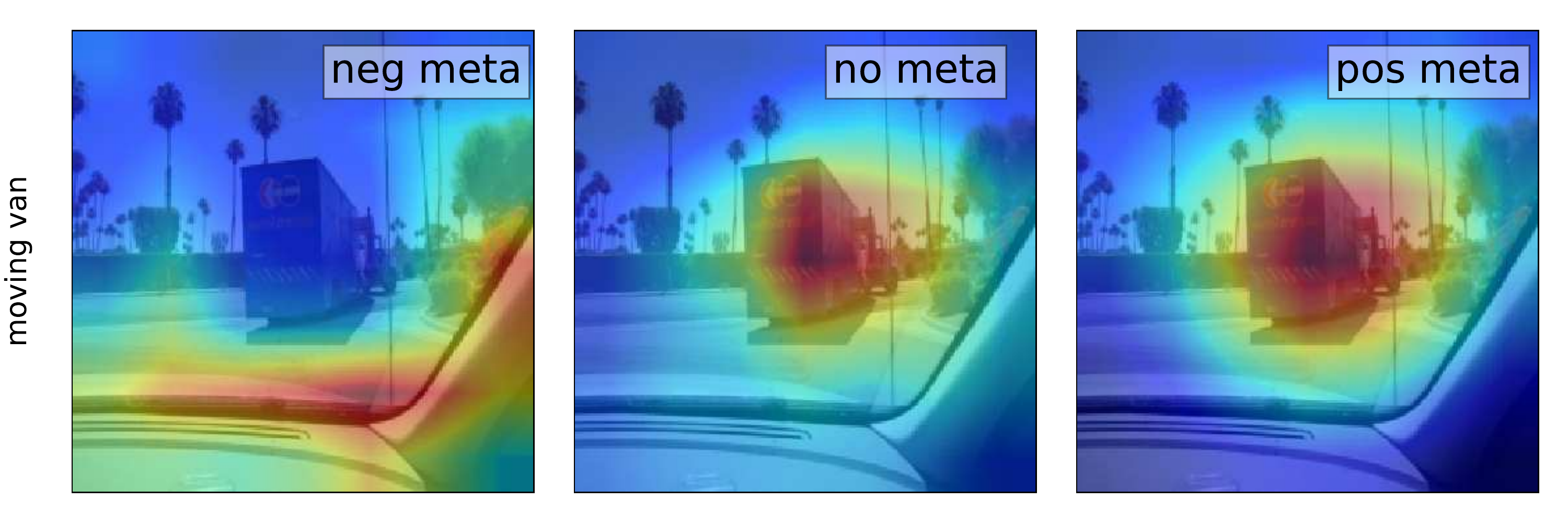}
    \includegraphics[width=\linewidth]{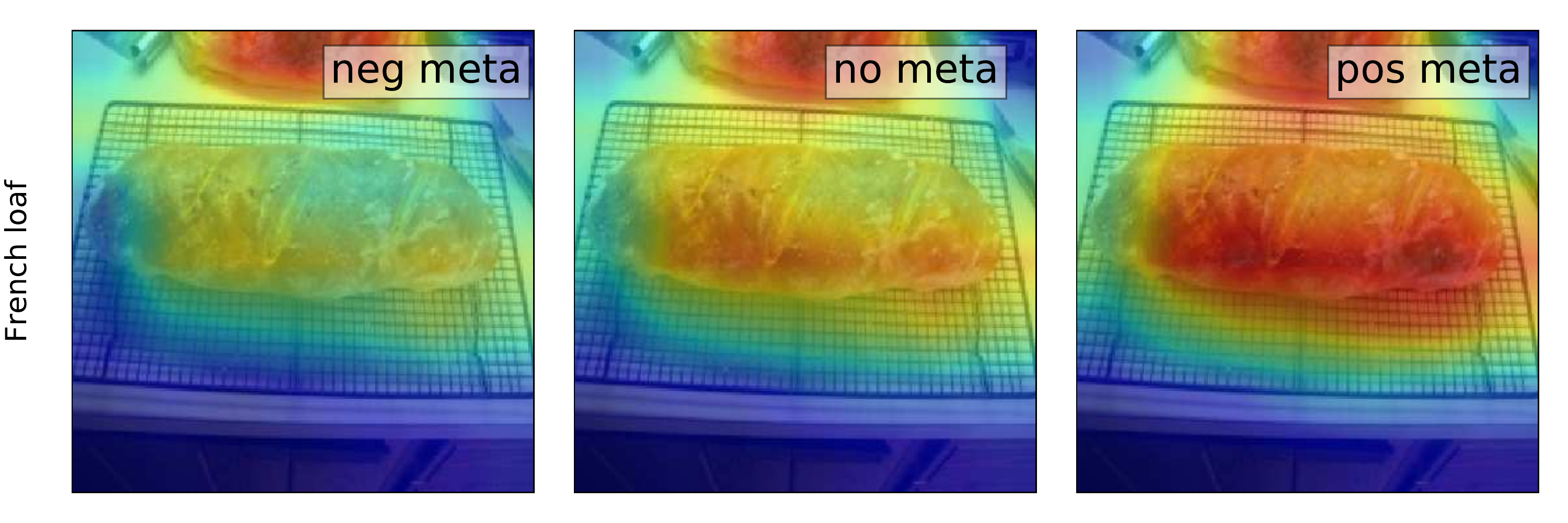}
    \caption{\textbf{Examples of positive and negative meta-saliency for selective NormGrad}. Negative meta-saliency (left images) corresponds to a gradient ascent inner step whereas a gradient descent step is used for the positive meta-saliency (images on the right). The center images do not use meta-saliency. } 
    \label{fig:pos_meta_neg_meta}
\end{figure}

\section{Model weights sensitivity}
\begin{figure*}[!htb]
    \centering
    \includegraphics[width=0.9\linewidth]{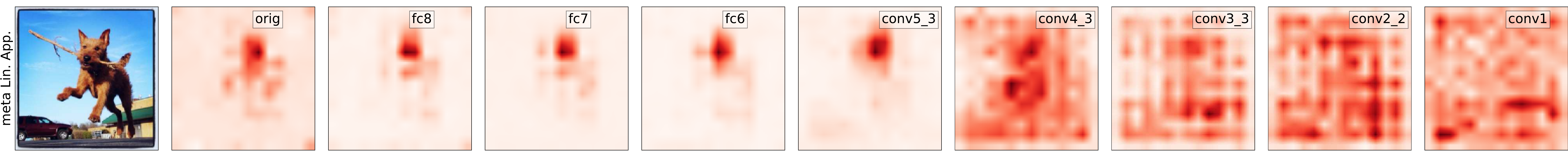}
    \includegraphics[width=0.9\linewidth]{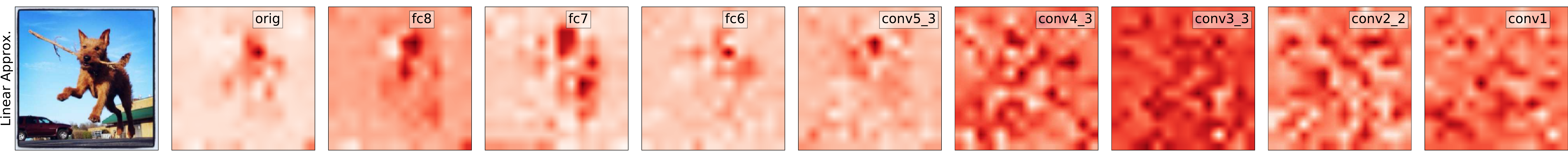}
    \includegraphics[width=0.9\linewidth]{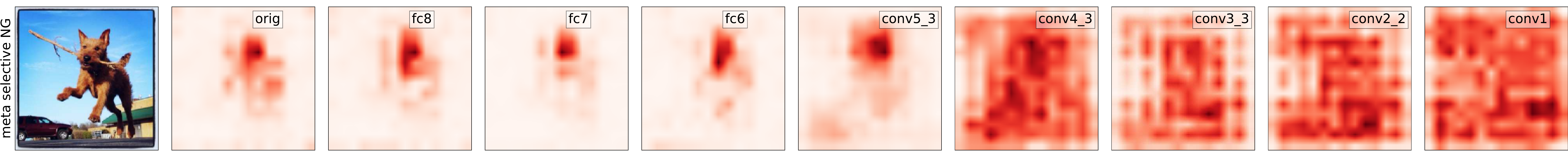}
    \includegraphics[width=0.9\linewidth]{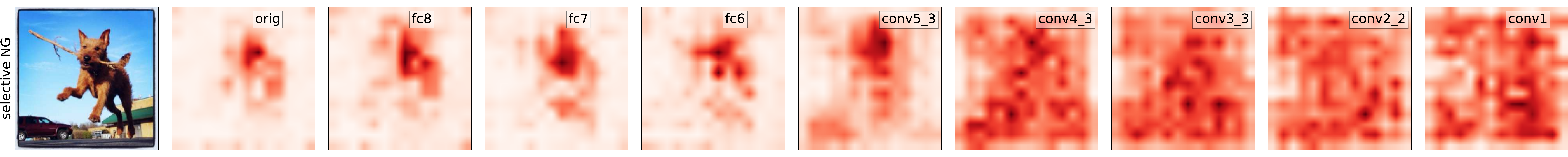}
    \includegraphics[width=0.9\linewidth]{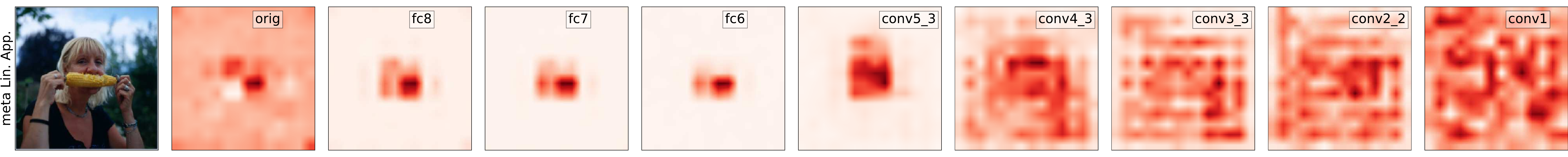}
    \includegraphics[width=0.9\linewidth]{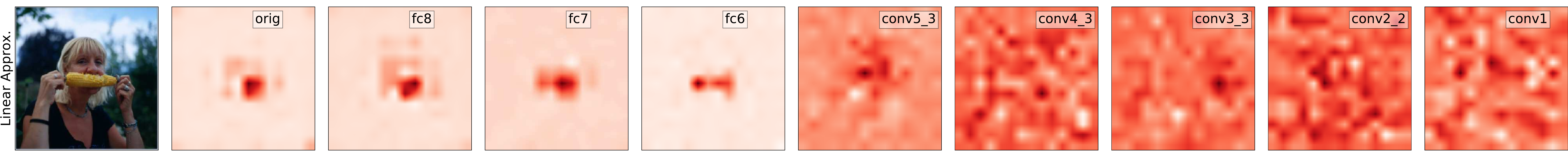}
    \includegraphics[width=0.9\linewidth]{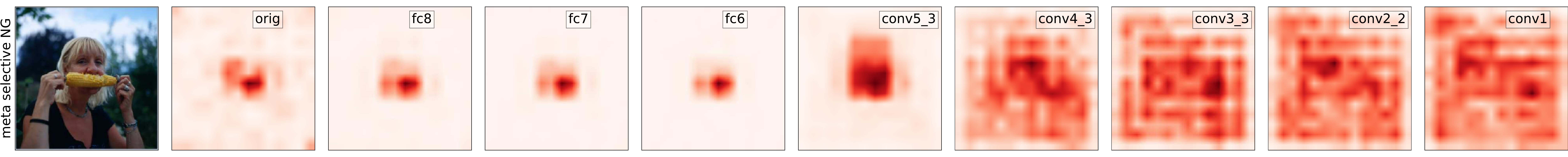}
    \includegraphics[width=0.9\linewidth]{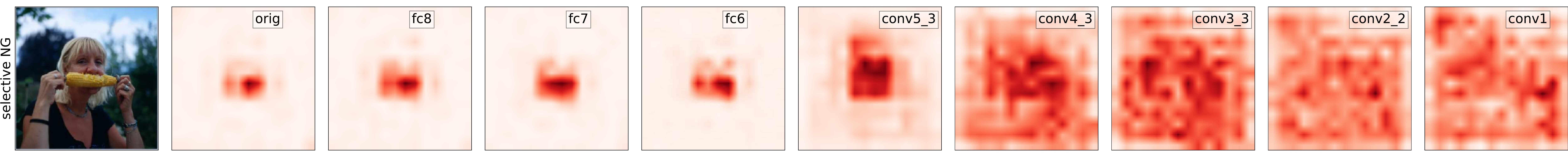}
    \includegraphics[width=0.9\linewidth]{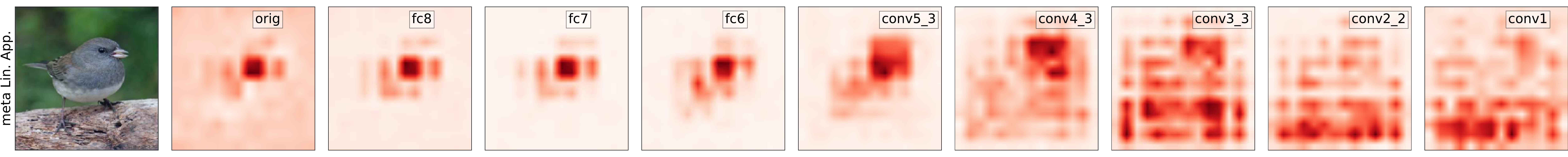}
    \includegraphics[width=0.9\linewidth]{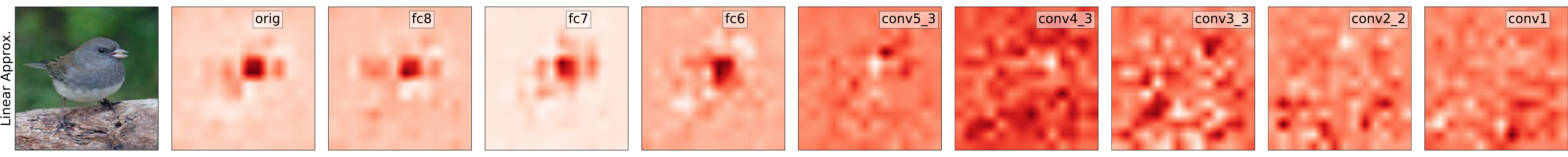}
    \includegraphics[width=0.9\linewidth]{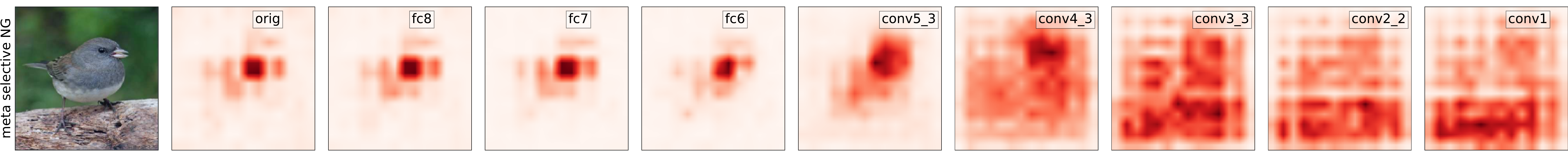}
    \includegraphics[width=0.9\linewidth]{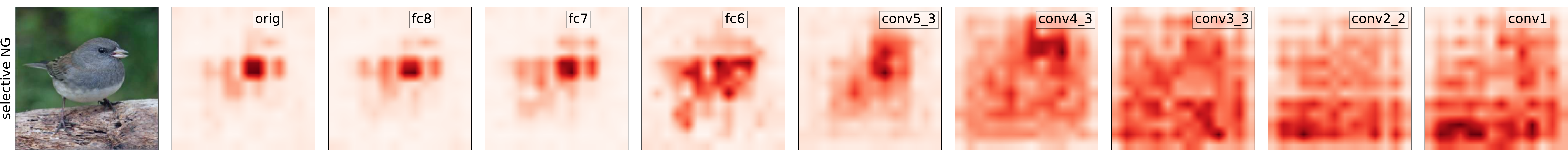}
    \caption{\textbf{Model weights sensitivity on VGG16.} Linear Approximation and selective NormGrad with and without meta-saliency.} 
    \label{fig:weight_sens}
\end{figure*}

In~\cref{fig:weight_sens}, we show the effects of cascading randomization on linear approximation and selective NormGrad  methods by using the same images as~\cite{Adebayo2018Sanity} (``junco'', ``corn'' and ``Irish terrier'') to illustrate the model weights sensitivity. 

Qualitatively, both methods - with and without meta-saliency - demonstrate the desired sensitivity to model weights as the saliency maps progressively lose focus from the object target as layer depth decreases. We also observe that using meta-saliency on top of the chosen base saliency method (shown on every other line of~\cref{fig:weight_sens}) delays the degradation in saliency maps to earlier layers.

\section{Image captioning}
\begin{figure*}[!h]
    \centering
    \includegraphics[width=\linewidth]{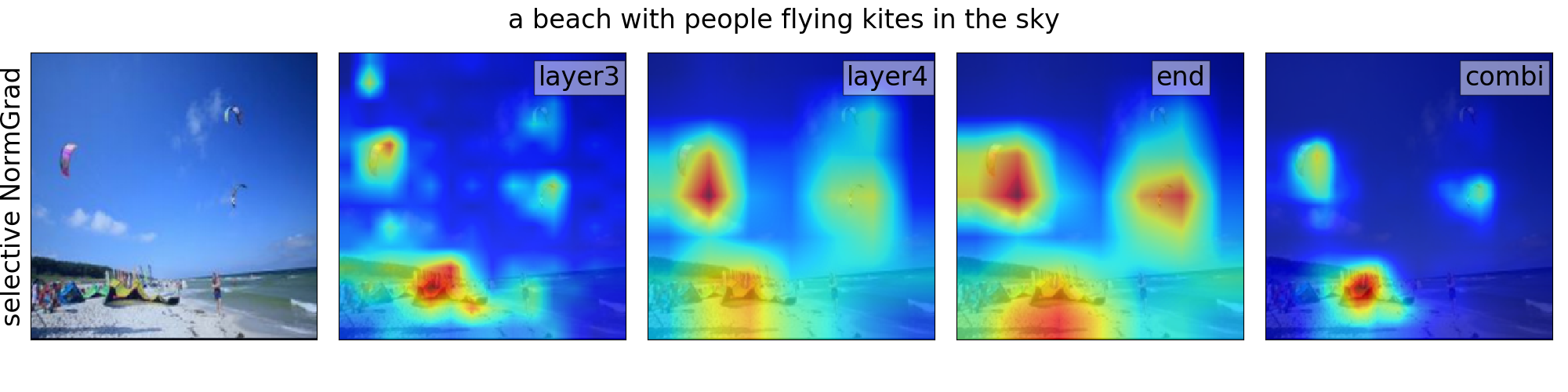}
    \includegraphics[width=\linewidth]{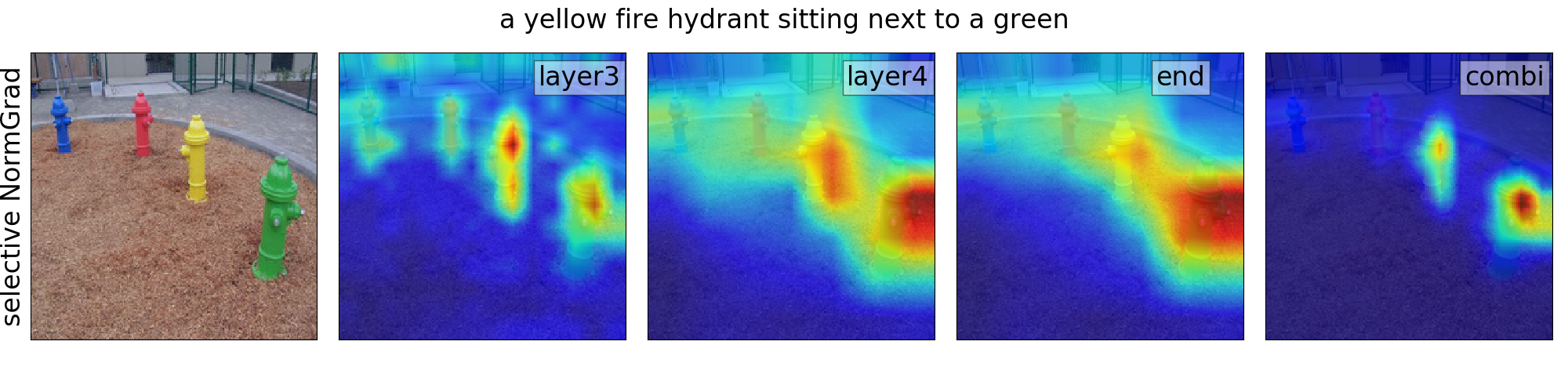}
    \includegraphics[width=\linewidth]{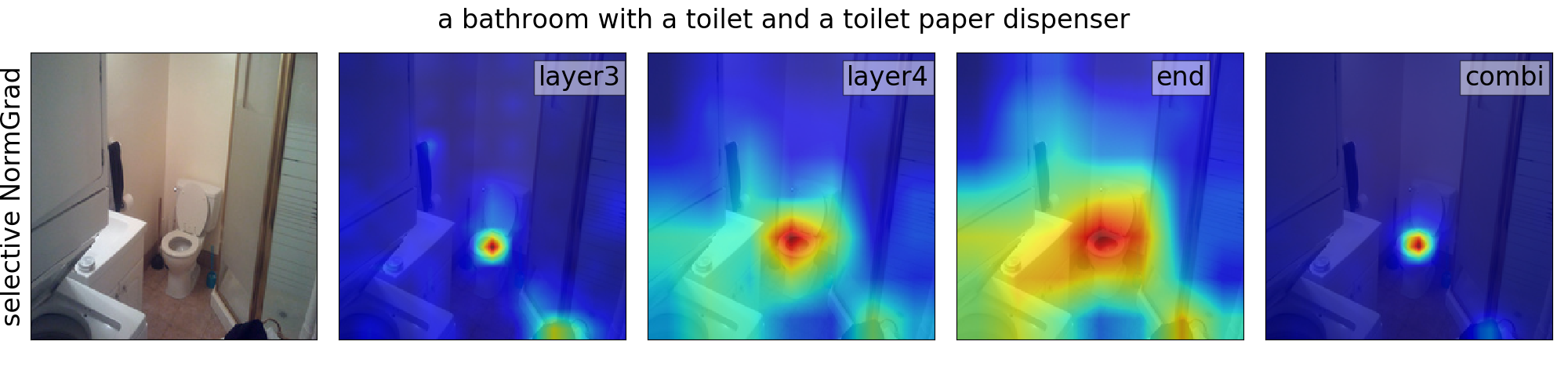}
    \caption{\textbf{Image captioning explanations.} Selective NormGrad is applied at different layers or combination of layers (last column) of a ResNet101. We observe that saliency maps at individual layers highlight a big part of the images. Using a combination of layers allows a clearer focus on the important parts of the images.} 
    \label{fig:captioning}
\end{figure*}

We show in this section a few examples of our proposed selective NormGrad method applied to the image captioning setting. We use a variant~\cite{luo2018discriminability} of the original neuraltalk2~\cite{karpathy2015deep} with a ResNet101 as backbone network followed by the LSTM caption model. As in~\cite{selvaraju17gradcam} we do a backward pass using the log probability of the generated caption as objective function. We apply selective NormGrad before the final global average pooling and at the ReLU layers just after the downsampling shortcuts of the third and fourth macro blocks. We also compare these saliency maps with the product combination map of these layers using a linear weighting scheme (see section 4.2 in the paper). We notice in~\cref{fig:captioning} that using a combination of layers produces a sharper saliency map than for individual layers.

\begin{figure*}[!h]
    \centering
    \includegraphics[width=\linewidth]{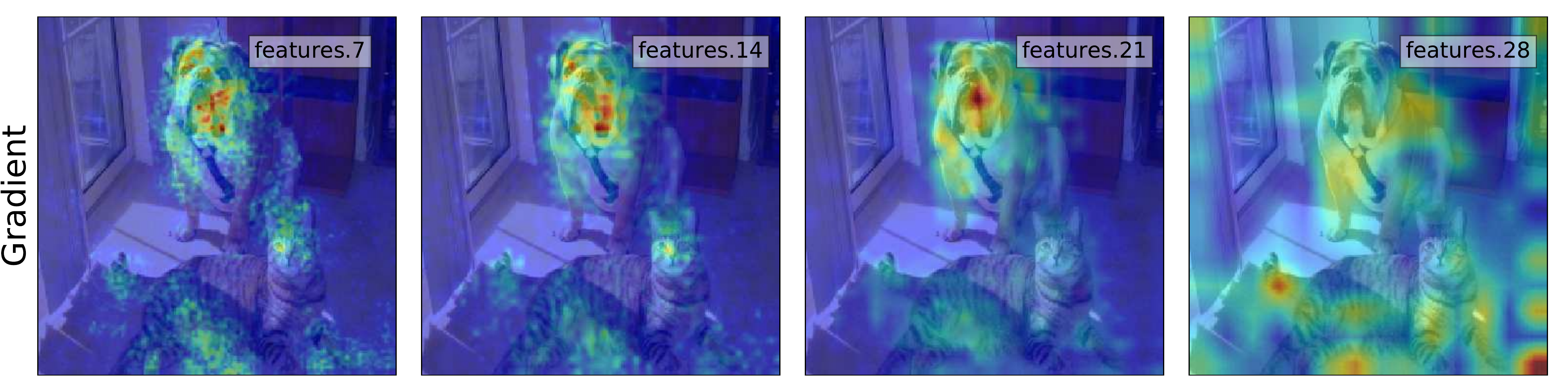}
    \includegraphics[width=\linewidth]{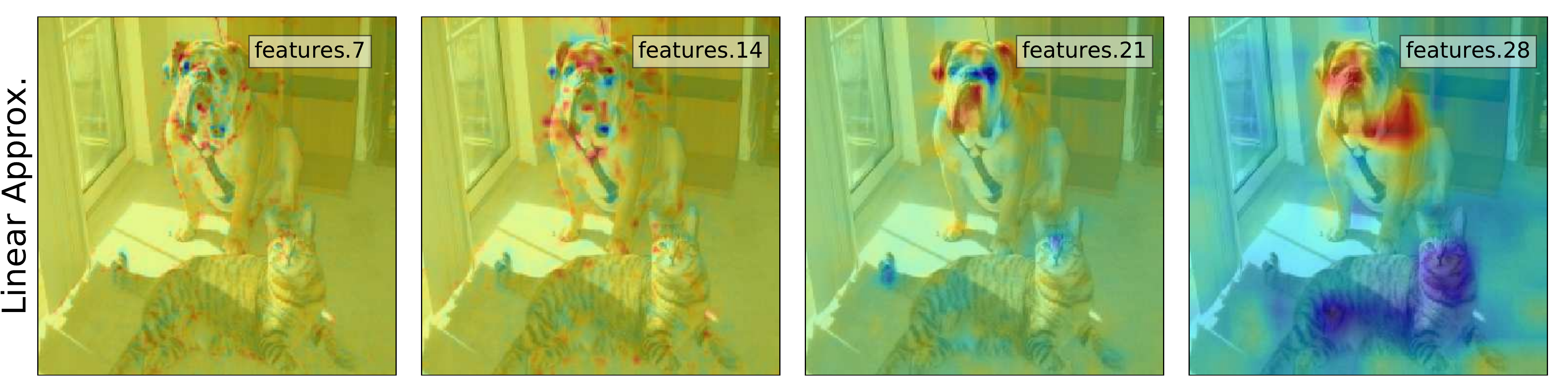}
    \includegraphics[width=\linewidth]{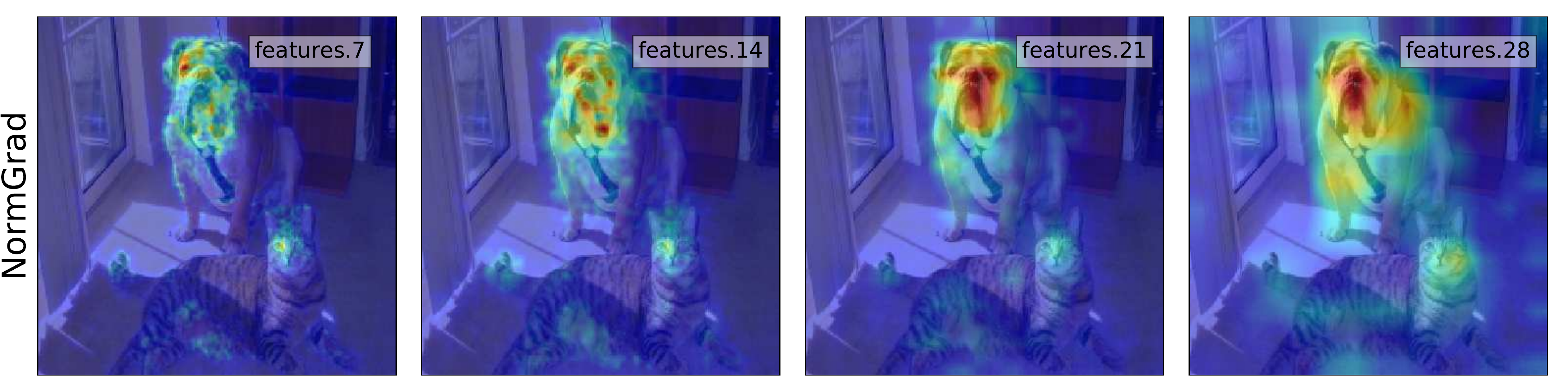}
    \includegraphics[width=\linewidth]{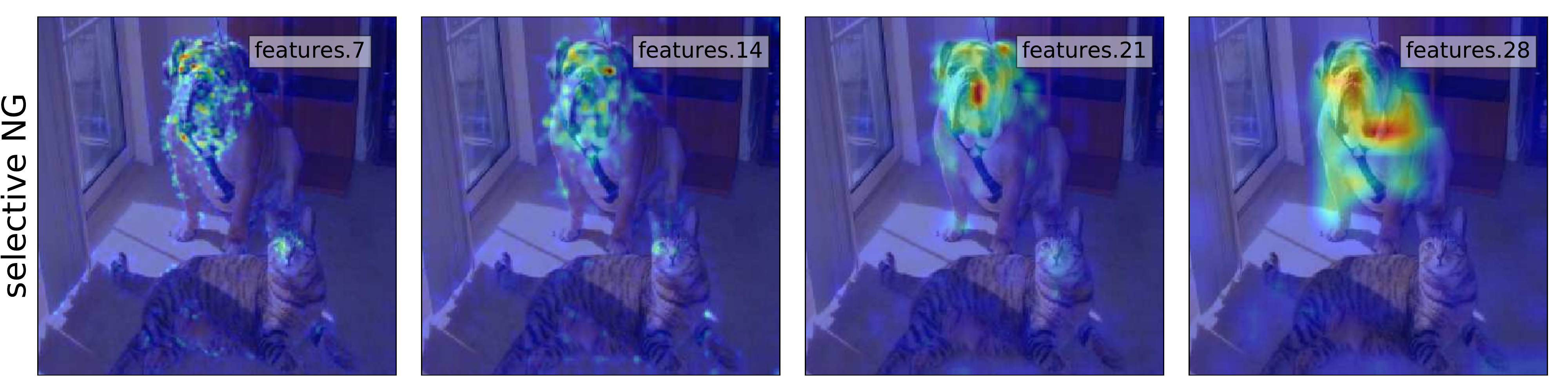}
    \caption{\textbf{Visualizations for some methods of the Extract \& Aggregate framework at different layers of VGG16.} The gradient backprogation method (first row) works well at all stages except at the end of the network. Selective NormGrad (last row), NormGrad (third row) and Linear Approximation (second row) perform well across the network. Finally we can observe that selective NormGrad and Linear Approximation are more class selective than NormGrad as the non targeted "tiger cat" appears more in the third row. } 
    \label{fig:framework_methods}
\end{figure*}